\documentclass[11pt]{article}

\usepackage[a4paper,margin=1in]{geometry}
\usepackage[T1]{fontenc}
\usepackage[utf8]{inputenc}
\usepackage{mathptmx}
\usepackage[scaled=0.9]{helvet}
\usepackage{microtype}

\usepackage{graphicx}
\usepackage{booktabs}
\usepackage{multirow}
\usepackage{array}
\usepackage[table]{xcolor}
\usepackage{float}
\usepackage{tabularx}
\usepackage{rotating}
\usepackage{caption}
\usepackage{subcaption}
\usepackage{enumitem}
\usepackage{amsmath,amssymb}
\usepackage{pifont}
\usepackage{tikz}
\usetikzlibrary{positioning,arrows.meta,calc,fit,backgrounds,shapes.misc,decorations.pathreplacing}
\usepackage[most]{tcolorbox}

\usepackage[numbers,sort&compress]{natbib}
\usepackage[colorlinks=true,linkcolor=ohBlue,citecolor=ohBlue,urlcolor=ohBlue]{hyperref}
\usepackage[capitalise,noabbrev]{cleveref}

\definecolor{ohBlue}{HTML}{1F4E9A}
\definecolor{catP}{HTML}{2F6FB5}   
\definecolor{catR}{HTML}{0E9A6E}   
\definecolor{catT}{HTML}{C47F12}   
\definecolor{catS}{HTML}{B23A5B}   
\definecolor{ink}{HTML}{222222}
\definecolor{muted}{HTML}{6B6B6B}
\definecolor{rule}{HTML}{D9D9D9}
\definecolor{postfill}{HTML}{E3E7EE}\definecolor{postdraw}{HTML}{9AA3AF}
\definecolor{bluefill}{HTML}{CCDAF5}\definecolor{bluedraw}{HTML}{4A6FB5}
\definecolor{greenfill}{HTML}{D8EAD3}\definecolor{greendraw}{HTML}{6AA35C}
\definecolor{orangefill}{HTML}{FFE8CC}\definecolor{orangedraw}{HTML}{E0A040}
\definecolor{redfill}{HTML}{F6D3D6}\definecolor{reddraw}{HTML}{C0566A}
\definecolor{itemfill}{HTML}{E6E6E6}\definecolor{itemdraw}{HTML}{666666}
 
\newif\ifshowplaceholders
\showplaceholderstrue

\newcommand{\bench}{\textsc{OpenHome}}
\newcommand{\cat}[2]{#2}

\setlist{itemsep=2pt,topsep=4pt}
\renewcommand{\arraystretch}{1.15}

\title{%
  \vspace{-1.2em}
  {\Large\bfseries \bench: From Perception to Insight\\[2pt] in Smart-Home Camera Videos}\\[8pt]
  {\large Technical Report of the Wyze OpenHome Benchmark}
}
\author{%
  Xianglong Wang$^{*}$\quad Neil Rodrigues$^{*}$\quad Pei Guo\quad Yiru Fang\quad Kevin Beussman\quad Shuyang Wang\\[2pt]
  Jingzhi Wang\quad Liangju Li\quad Shunyao Xu\quad Zhongwei Cheng\quad Wei Li$^{\dagger}$\quad Lin Chen\\[6pt]
  {\small Wyze Labs}\\[2pt]
  {\small Project page: \url{https://huggingface.co/datasets/wyzelabs/openhome}}\\[4pt]
  {\small $^{*}$Equal contribution.\quad $^{\dagger}$Corresponding author: \href{mailto:wei.li@wyze.com}{wei.li@wyze.com}}
}
\date{October 2026}

\begin{document}
\maketitle

\begin{abstract}
\noindent
Vision language models (VLMs) have been increasingly adopted in consumer smart-home camera systems to interpret recorded videos and provide insights. However, video clips captured from smart-home cameras are different from curated web videos dominating existing VLM benchmarks. Smart-home video clips often do not contain subtitles or narration and are captured from different viewing angles, under various lighting conditions, and in diverse scenarios. However, the capabilities of the model in this domain, under sparse visual signals and limited context, are not fully assessed by existing VLM benchmarks. To close this gap, we introduce \bench{}, a visual-only, four-option multiple-choice benchmark for evaluating smart-home video understanding. It contains 2,254 questions on 1,118 real-world video clips and organizes 10 tasks into three capability levels. 
\cat{catP}{Perception} evaluates visible content, \cat{catR}{Reasoning} evaluates causal, spatial, and temporal relationships, and
\cat{catS}{Insight} evaluates subject intent, supporting evidence and appropriate homeowner responses. 
Based on the dataset, we describe the task design, data pipeline and evaluation
protocol, and report results for 32 closed-source and open-weight VLMs. The best model, Gemini 4
Argon, reaches 81.2\% overall accuracy. Frontier closed-source models outperform open-weight models on every
level. The best closed-source model leads the best open-weight model by 16.3\% on
Perception, 22.4\% on Reasoning and 37.4\% on Insight. 
\end{abstract}


\section{Introduction}
\label{sec:intro}

Smart home cameras are installed in millions of homes. Their value increasingly extends beyond recording footage to helping homeowners understand, quickly and accurately, \emph{what happened and whether it matters}: a package is taken from the porch, a wild animal approaches the door, or a visitor rings the bell and waits outside. Vision language models (VLMs) offer a way to interpret these events, and recent benchmarks show steady progress in general video understanding~\citep{fu2025videomme,li2024mvbench,zhou2024mlvu,wu2024longvideobench}.

However, smart home footage presents viewing conditions and contextual limitations that general video benchmarks do not systematically assess. Cameras are positioned to monitor a particular area, and their placement shapes what is visible: a camera mounted high above a doorway may show a person mainly from above, while one placed near the ground captures the scene from a low angle. Nearby walls, doorframes, or vegetation can partially obstruct the view, leaving relevant actions or objects hidden. Wide angle distortion, low light and infrared recording can further obscure visual details. Coverage is also limited in time. Recording is often triggered by an event and may begin after an action has started or end before it finishes, leaving only a fragment of the event visible. With no accompanying narration and little context beyond the clip, a model must interpret what happened from incomplete visual evidence while recognizing what the footage cannot establish. Under these conditions, limitations in \emph{what} a model perceives, or \emph{when} it identifies a particular event, may constrain its interpretation of the scene.

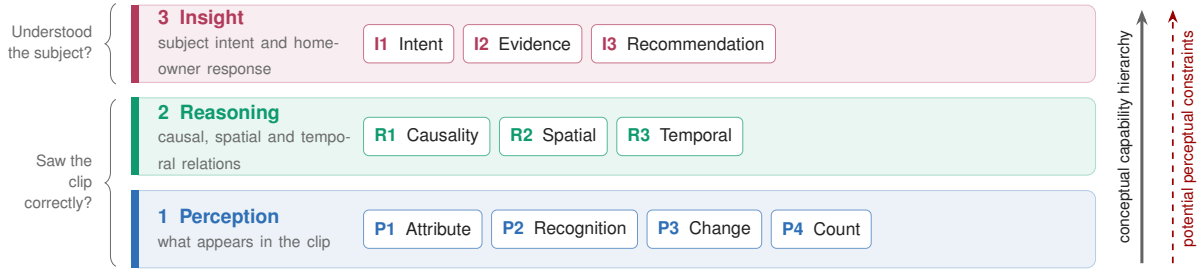
\begin{figure}[t]
\centering
\resizebox{\textwidth}{!}{%
\begin{tikzpicture}[
  font=\sffamily\small,
  band/.style={rounded corners=4pt, minimum height=1.25cm, minimum width=15.6cm, inner sep=0},
  head/.style={align=left, text width=3.15cm, anchor=west, font=\sffamily\small},
  chip/.style={rounded corners=2.5pt, draw=#1!70, fill=white, text=ink,
               minimum height=0.62cm, inner xsep=5pt, font=\sffamily\footnotesize, anchor=west},
]
  \foreach \y/\c in {0/catP, 1.5/catR, 3.0/catS}{
    \node[band, fill=\c!9, draw=\c!35] at (7.8,\y) {};
    \fill[\c] (0.0,\y-0.625) rectangle ++(0.12,1.25);
  }
  \node[head] at (0.3,0)   {\textcolor{catP}{\textbf{1\;\;Perception}}\\[-1pt]\textcolor{muted}{\scriptsize what appears in the clip}};
  \node[head] at (0.3,1.5) {\textcolor{catR}{\textbf{2\;\;Reasoning}}\\[-1pt]\textcolor{muted}{\scriptsize causal, spatial and temporal relations}};
  \node[head] at (0.3,3.0) {\textcolor{catS}{\textbf{3\;\;Insight}}\\[-1pt]\textcolor{muted}{\scriptsize subject intent and homeowner response}};

  \node[chip=catP] (p1) at (3.75,0) {\textbf{\textcolor{catP}{P1}}\; Attribute};
  \node[chip=catP, right=4pt of p1] (p2) {\textbf{\textcolor{catP}{P2}}\; Recognition};
  \node[chip=catP, right=4pt of p2] (p3) {\textbf{\textcolor{catP}{P3}}\; Change};
  \node[chip=catP, right=4pt of p3] (p4) {\textbf{\textcolor{catP}{P4}}\; Count};
  \node[chip=catR] (r1) at (3.75,1.5) {\textbf{\textcolor{catR}{R1}}\; Causality};
  \node[chip=catR, right=4pt of r1] (r2) {\textbf{\textcolor{catR}{R2}}\; Spatial};
  \node[chip=catR, right=4pt of r2] (r3) {\textbf{\textcolor{catR}{R3}}\; Temporal};
  \node[chip=catS] (i1) at (3.75,3.0) {\textbf{\textcolor{catS}{I1}}\; Intent};
  \node[chip=catS, right=4pt of i1] (i2) {\textbf{\textcolor{catS}{I2}}\; Evidence};
  \node[chip=catS, right=4pt of i2] (i3) {\textbf{\textcolor{catS}{I3}}\; Recommendation};

  \draw[-{Stealth[length=7pt]}, line width=1.4pt, ink!70] (16.35,-0.55) -- (16.35,3.55)
    node[midway, rotate=90, above, font=\sffamily\scriptsize, text=ink] {conceptual capability hierarchy};
  \draw[-{Stealth[length=6pt]}, line width=0.8pt, red!60!black, dashed] (16.85,-0.55) -- (16.85,3.55)
    node[midway, rotate=90, below=1pt, font=\sffamily\scriptsize, text=red!60!black] {potential perceptual constraints};

  \draw[decorate, decoration={brace, amplitude=5pt, mirror}, muted] (-0.25,2.1) -- (-0.25,-0.6)
    node[midway, left=7pt, align=right, font=\sffamily\scriptsize, text=muted] {Saw the\\clip\\correctly?};
  \draw[decorate, decoration={brace, amplitude=5pt, mirror}, muted] (-0.25,3.65) -- (-0.25,2.35)
    node[midway, left=7pt, align=right, font=\sffamily\scriptsize, text=muted] {Understood\\the subject?};
\end{tikzpicture}}
\caption{\textbf{The \bench{} capability ladder.} Three capability levels and 10 tasks, ordered from
Perception to Insight. Perception and Reasoning check that a model saw and followed the clip
correctly; Insight checks that it understood the subject and knows what the homeowner should do.
Tasks are scored independently. The arrows indicate hypothesised relationships, not measured error propagation.}
\label{fig:teaser}
\end{figure}

We introduce \textbf{\bench}, a benchmark designed to assess video understanding exactly under the above-mentioned constraints. Camera viewpoints and partial visibility challenge a model's ability to identify subjects, attributes and changes in the scene. Even when these details are recognized, gaps in the recording can make it difficult to establish how events relate in space, time and cause. Interpreting a subject's intent and determining an appropriate homeowner response requires a further judgement about what the available evidence supports. To examine these aspects of understanding, \bench{} organizes 10 tasks into three capability levels (\cref{fig:teaser}): \cat{catP}{Perception} assesses what is visibly present and what changes; \cat{catR}{Reasoning} assesses spatial, temporal and causal relationships; and \cat{catS}{Insight} assesses subject intent, supporting evidence and appropriate homeowner responses. Each level is scored separately to provide a complete capability profile.

\paragraph{Contributions.}
\begin{itemize}[leftmargin=1.4em]
  \item \textbf{A benchmark grounded in real smart home footage.} \bench{} comprises 2,254 questions on 1,118 camera clips covering people, vehicles, animals and natural phenomena. It evaluates video understanding under practical viewing conditions, including restricted visibility, challenging lighting and incomplete recordings of events.

  \item \textbf{A capability framework connecting visual understanding to homeowner decisions.} We organise 10 tasks into Perception, Reasoning and Insight, covering visible content, event relationships, subject intent, supporting evidence and appropriate homeowner responses. The three Insight tasks are independently posed and scored, so each assesses a distinct aspect of interpretation without relying on a model's answers to the other probes. Together, the tasks provide a detailed capability profile beyond overall accuracy.

  \item \textbf{An empirical assessment of 32 vision language models.} Under the evaluated input and reasoning settings, counting and temporal ordering remain challenging even for the strongest model. The gap between the best closed models and models with open weights is largest on Insight. These findings highlight differences across capabilities that a single aggregate score would obscure.
\end{itemize}

\section{Related Work}
\label{sec:related}

\paragraph{General video understanding benchmarks.}
Early video question answering (QA) datasets such as ActivityNet-QA~\citep{yu2019activitynetqa} and
NExT-QA~\citep{xiao2021nextqa} evaluated description, temporal and causal questions on web and
everyday videos. With the rise of VLMs, comprehensive suites such as
MVBench~\citep{li2024mvbench} and Video-MME~\citep{fu2025videomme} broadened task coverage and
video durations, while MLVU~\citep{zhou2024mlvu}, LongVideoBench~\citep{wu2024longvideobench} and
EgoSchema~\citep{mangalam2023egoschema} target long-form understanding. These benchmarks draw
predominantly on edited, well-lit, often narrated content; \bench{} instead targets short,
unedited clips from fixed smart home cameras where the signal is sparse.

\paragraph{Diagnostic, temporal and causal evaluation.}
Perception Test~\citep{patraucean2023perception} and TempCompass~\citep{liu2024tempcompass}
isolate perceptual and temporal skills; CLEVRER~\citep{yi2020clevrer} studies causal reasoning in
synthetic scenes. \bench{} adopts the diagnostic spirit of these works but grounds every task in
the question a homeowner would actually ask.

\paragraph{Social and intent understanding.}
Social-IQ~\citep{zadeh2019socialiq} evaluates understanding of human intent and social
situations in conversational videos. Video-MME-v2~\citep{fu2026videommev2} places social
dynamics at the top of a three-level hierarchy, from visual information aggregation through
temporal dynamics to complex reasoning, and includes inferring an individual's intent, emotion
or mental state from multimodal evidence. It also reports that errors in visual aggregation and
temporal modelling propagate into higher-level reasoning. This motivates the \bench{} capability
framework, which examines how models interpret behaviour in unedited smart-home camera footage using visual evidence alone.

\paragraph{Surveillance and home video.}
Surveillance datasets such as UCF-Crime~\citep{sultani2018ucfcrime} frame the problem as anomaly
detection, and Charades~\citep{sigurdsson2016charades} records scripted indoor activities.
SmartHome Bench~\citep{zhao2025smarthomebench} focuses specifically on anomaly detection in smart home camera videos, evaluating whether multimodal large language models can identify anomalous events and explain them. \bench{} instead evaluates a broader range of video-understanding capabilities in this setting, including visible content and changes, relationships, and subject intent. It further assesses whether models can ground their interpretations
in visual evidence and recommend appropriate homeowner responses, including when no action is warranted.

\section{The \bench{} Benchmark}
\label{sec:benchmark}

\subsection{Design principles}
\label{sec:principles}

\begin{enumerate}[leftmargin=1.6em, label=\textbf{(\arabic*)}]
  \item \textbf{Hierarchical capability assessment.} Higher-level judgements may depend on lower-level
        perception. Measuring each level separately provides a capability profile, across different levels of video understanding.
  \item \textbf{Visual grounding across the clip.} Questions may depend on any moment of the clip
        and on any visual cue---movement, gaze, posture, objects handled. Every question,
        including every Insight probe, is answerable from the frames alone, without audio.
  \item \textbf{Behaviour-based assessment.} All questions are judged solely on what a
        subject \emph{does}. Annotation guidelines prohibit answers that depend on a subject's
        appearance, demographic attributes or identity.
  \item \textbf{Independent probes.} Each probe is a standalone question that
        supplies the context required to answer it. A wrong answer on one probe never carries into the
        next, so every probe can be scored on its own.
  \item \textbf{Homeowner-centred answers.} Correct answers reflect what a homeowner
        can reasonably conclude and do, including the option that nothing needs to be done.
\end{enumerate}

\subsection{Task taxonomy}
\label{sec:taxonomy}

\bench{} has 10 tasks in three capability levels: Perception (P1--P4) asks what appears in the clip, Reasoning
(R1--R3) assesses causal, spatial and temporal relationships, and Insight (I1--I3) assesses subject intent and appropriate homeowner responses.
\Cref{tab:taxonomy} provides an overview of the three levels and defines each task.

\begin{table}[!t]
\centering
\small
\caption{\textbf{\bench{} task taxonomy.} Three capability levels, 10 tasks.}
\label{tab:taxonomy}
\begin{tabularx}{\textwidth}{@{}>{\bfseries}l l X@{}}
\toprule
\textnormal{\textbf{Code}} & \textbf{Task} & \textbf{What it tests} \\
\midrule
\multicolumn{3}{@{}l}{\textbf{1\;\;Perception} --- \emph{what appears in the clip}}\\
{P1} & Attribute & Describe visible properties of people, vehicles, animals and objects through the clip, e.g.\ colour, clothing, size. \\
{P2} & Recognition & Identify the category or type of a visible subject, e.g.\ a bear, a fox, a child, a delivery truck. \\
{P3} & Change & Identify changes between the beginning and end of the clip, e.g.\ a package disappearing, a door left open, a car changing position. \\
{P4} & Count & Count people or objects while accounting for the same subject leaving and re-entering the frame. \\
\midrule
\multicolumn{3}{@{}l}{\textbf{2\;\;Reasoning} --- \emph{causal, spatial and temporal relationships}}\\
{R1} & Causality & Determine the actual cause of an event, e.g.\ a trash can falls as a person walks past, but the wind, not the person, knocked it over. \\
{R2} & Spatial & Describe the relative positions of entities in the scene, e.g.\ up or down, left or right, in or out. \\
{R3} & Temporal & Determine the order of events or movements in a clip, e.g.\ what was done first, and what came next. \\
\midrule
\multicolumn{3}{@{}l}{\textbf{3\;\;Insight} --- \emph{subject intent and homeowner response}}\\
{I1} & Intent & Infer a subject's primary goal, from visual cues such as movement, gaze, posture, objects handled. \\
{I2} & Evidence & Identify visual evidence that supports a proposed intent, or determine which visual cue does not. \\
{I3} & Recommendation & Recommend an appropriate homeowner response based on observed events, e.g.\ taking no action, reporting the incident, making repairs. \\
\bottomrule
\end{tabularx}
\end{table}

\begin{figure}[p]
\centering
\resizebox{\textwidth}{!}{%
\begin{tikzpicture}
\begin{scope}[shift={(0.00,0.00)}]
\node[anchor=north west, inner sep=0, font=\sffamily\fontsize{9}{10.5}\selectfont\bfseries, text=catP] at (0,0) {(a) Perception: P1 Attribute};
\node[anchor=north west, inner sep=0] (fa0) at (0.00,-0.43) {\includegraphics[width=2.5cm,height=1.40625cm]{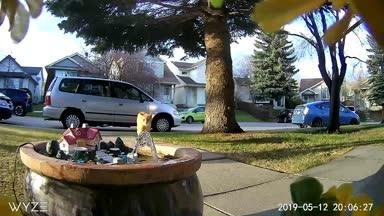}};
\node[anchor=north west, fill=black!70, text=white, font=\sffamily\fontsize{7}{8}\selectfont, inner sep=1.4pt] at (fa0.north west) {0:01};
\node[anchor=north, text width=2.5cm, align=center, inner sep=0, font=\sffamily\fontsize{7.3}{8.4}\selectfont\itshape, text=ink] at ($(fa0.south)+(0,-0.09)$) {Cat walks up the path};
\node[anchor=north west, inner sep=0] (fa1) at (2.60,-0.43) {\includegraphics[width=2.5cm,height=1.40625cm]{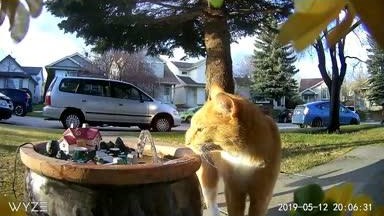}};
\node[anchor=north west, fill=black!70, text=white, font=\sffamily\fontsize{7}{8}\selectfont, inner sep=1.4pt] at (fa1.north west) {0:05};
\node[anchor=north, text width=2.5cm, align=center, inner sep=0, font=\sffamily\fontsize{7.3}{8.4}\selectfont\itshape, text=ink] at ($(fa1.south)+(0,-0.09)$) {Reaches the fountain};
\node[anchor=north west, inner sep=0] (fa2) at (5.20,-0.43) {\includegraphics[width=2.5cm,height=1.40625cm]{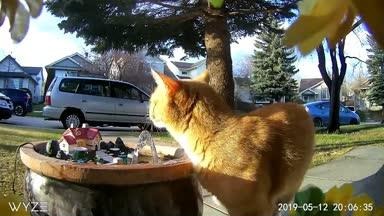}};
\node[anchor=north west, fill=black!70, text=white, font=\sffamily\fontsize{7}{8}\selectfont, inner sep=1.4pt] at (fa2.north west) {0:09};
\node[anchor=north, text width=2.5cm, align=center, inner sep=0, font=\sffamily\fontsize{7.3}{8.4}\selectfont\itshape, text=ink] at ($(fa2.south)+(0,-0.09)$) {Drinks from the spout};
\node[anchor=north west, draw=catP!35, fill=catP!4, rounded corners=2pt, text width=7.35cm, minimum height=1.05cm, inner sep=4.5pt, align=left, font=\sffamily\fontsize{8.5}{10}\selectfont] at (0,-2.56) {What colour is the cat in this video?\\[3pt]\textcolor{ink}{\ding{51}\;Orange with a white chest.}};
\end{scope}

\begin{scope}[shift={(8.15,0.00)}]
\node[anchor=north west, inner sep=0, font=\sffamily\fontsize{9}{10.5}\selectfont\bfseries, text=catP] at (0,0) {(b) Perception: P2 Recognition};
\node[anchor=north west, inner sep=0] (fb0) at (0.00,-0.43) {\includegraphics[width=2.5cm,height=1.40625cm]{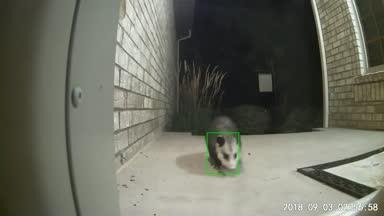}};
\node[anchor=north west, fill=black!70, text=white, font=\sffamily\fontsize{7}{8}\selectfont, inner sep=1.4pt] at (fb0.north west) {0:01};
\node[anchor=north, text width=2.5cm, align=center, inner sep=0, font=\sffamily\fontsize{7.3}{8.4}\selectfont\itshape, text=ink] at ($(fb0.south)+(0,-0.09)$) {Climbs onto the porch};
\node[anchor=north west, inner sep=0] (fb1) at (2.60,-0.43) {\includegraphics[width=2.5cm,height=1.40625cm]{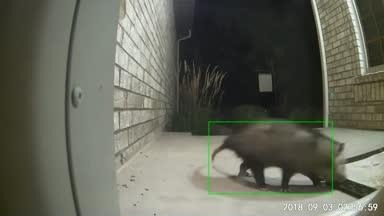}};
\node[anchor=north west, fill=black!70, text=white, font=\sffamily\fontsize{7}{8}\selectfont, inner sep=1.4pt] at (fb1.north west) {0:02};
\node[anchor=north, text width=2.5cm, align=center, inner sep=0, font=\sffamily\fontsize{7.3}{8.4}\selectfont\itshape, text=ink] at ($(fb1.south)+(0,-0.09)$) {Crosses the doormat};
\node[anchor=north west, inner sep=0] (fb2) at (5.20,-0.43) {\includegraphics[width=2.5cm,height=1.40625cm]{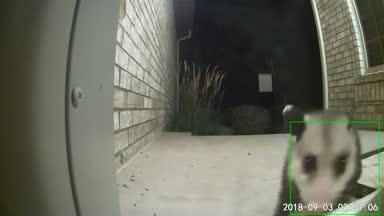}};
\node[anchor=north west, fill=black!70, text=white, font=\sffamily\fontsize{7}{8}\selectfont, inner sep=1.4pt] at (fb2.north west) {0:09};
\node[anchor=north, text width=2.5cm, align=center, inner sep=0, font=\sffamily\fontsize{7.3}{8.4}\selectfont\itshape, text=ink] at ($(fb2.south)+(0,-0.09)$) {Returns to sniff the camera};
\node[anchor=north west, draw=catP!35, fill=catP!4, rounded corners=2pt, text width=7.35cm, minimum height=1.05cm, inner sep=4.5pt, align=left, font=\sffamily\fontsize{8.5}{10}\selectfont] at (0,-2.56) {Which animal appears in this video?\\[3pt]\textcolor{ink}{\ding{51}\;An opossum.}};
\end{scope}

\begin{scope}[shift={(0.00,-4.15)}]
\node[anchor=north west, inner sep=0, font=\sffamily\fontsize{9}{10.5}\selectfont\bfseries, text=catP] at (0,0) {(c) Perception: P3 Change};
\node[anchor=north west, inner sep=0] (fc0) at (0.00,-0.43) {\includegraphics[width=2.5cm,height=1.40625cm]{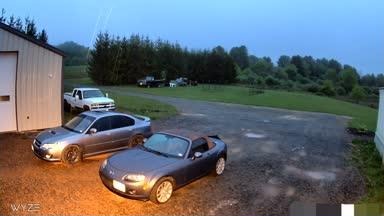}};
\node[anchor=north west, fill=black!70, text=white, font=\sffamily\fontsize{7}{8}\selectfont, inner sep=1.4pt] at (fc0.north west) {0:01};
\node[anchor=north, text width=2.5cm, align=center, inner sep=0, font=\sffamily\fontsize{7.3}{8.4}\selectfont\itshape, text=ink] at ($(fc0.south)+(0,-0.09)$) {Driver door is closed};
\node[anchor=north west, inner sep=0] (fc1) at (2.60,-0.43) {\includegraphics[width=2.5cm,height=1.40625cm]{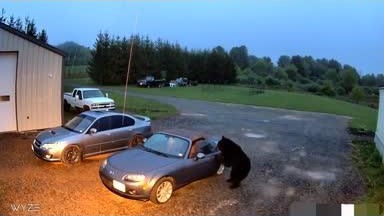}};
\node[anchor=north west, fill=black!70, text=white, font=\sffamily\fontsize{7}{8}\selectfont, inner sep=1.4pt] at (fc1.north west) {0:06};
\node[anchor=north, text width=2.5cm, align=center, inner sep=0, font=\sffamily\fontsize{7.3}{8.4}\selectfont\itshape, text=ink] at ($(fc1.south)+(0,-0.09)$) {Bear pulls the door open};
\node[anchor=north west, inner sep=0] (fc2) at (5.20,-0.43) {\includegraphics[width=2.5cm,height=1.40625cm]{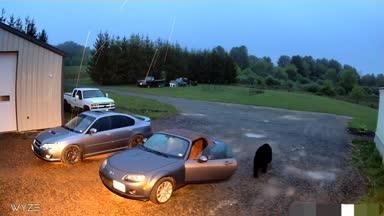}};
\node[anchor=north west, fill=black!70, text=white, font=\sffamily\fontsize{7}{8}\selectfont, inner sep=1.4pt] at (fc2.north west) {0:33};
\node[anchor=north, text width=2.5cm, align=center, inner sep=0, font=\sffamily\fontsize{7.3}{8.4}\selectfont\itshape, text=ink] at ($(fc2.south)+(0,-0.09)$) {Bear leaves; door stays open};
\node[anchor=north west, draw=catP!35, fill=catP!4, rounded corners=2pt, text width=7.35cm, minimum height=1.05cm, inner sep=4.5pt, align=left, font=\sffamily\fontsize{8.5}{10}\selectfont] at (0,-2.56) {How has the car changed by the end of the clip?\\[3pt]\textcolor{ink}{\ding{51}\;Its driver door is left open.}};
\end{scope}

\begin{scope}[shift={(8.15,-4.15)}]
\node[anchor=north west, inner sep=0, font=\sffamily\fontsize{9}{10.5}\selectfont\bfseries, text=catP] at (0,0) {(d) Perception: P4 Count};
\node[anchor=north west, inner sep=0] (fd0) at (0.00,-0.43) {\includegraphics[width=2.5cm,height=1.40625cm]{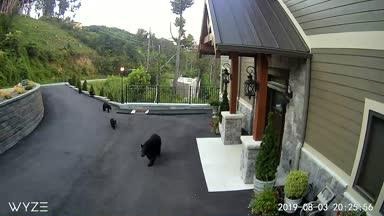}};
\node[anchor=north west, fill=black!70, text=white, font=\sffamily\fontsize{7}{8}\selectfont, inner sep=1.4pt] at (fd0.north west) {0:00};
\node[anchor=north, text width=2.5cm, align=center, inner sep=0, font=\sffamily\fontsize{7.3}{8.4}\selectfont\itshape, text=ink] at ($(fd0.south)+(0,-0.09)$) {Adult bear leads the way};
\node[anchor=north west, inner sep=0] (fd1) at (2.60,-0.43) {\includegraphics[width=2.5cm,height=1.40625cm]{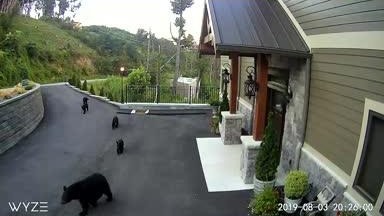}};
\node[anchor=north west, fill=black!70, text=white, font=\sffamily\fontsize{7}{8}\selectfont, inner sep=1.4pt] at (fd1.north west) {0:04};
\node[anchor=north, text width=2.5cm, align=center, inner sep=0, font=\sffamily\fontsize{7.3}{8.4}\selectfont\itshape, text=ink] at ($(fd1.south)+(0,-0.09)$) {Cubs follow in a line};
\node[anchor=north west, inner sep=0] (fd2) at (5.20,-0.43) {\includegraphics[width=2.5cm,height=1.40625cm]{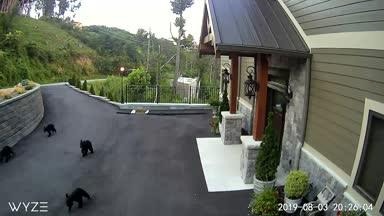}};
\node[anchor=north west, fill=black!70, text=white, font=\sffamily\fontsize{7}{8}\selectfont, inner sep=1.4pt] at (fd2.north west) {0:08};
\node[anchor=north, text width=2.5cm, align=center, inner sep=0, font=\sffamily\fontsize{7.3}{8.4}\selectfont\itshape, text=ink] at ($(fd2.south)+(0,-0.09)$) {Last cubs leave the frame};
\node[anchor=north west, draw=catP!35, fill=catP!4, rounded corners=2pt, text width=7.35cm, minimum height=1.05cm, inner sep=4.5pt, align=left, font=\sffamily\fontsize{8.5}{10}\selectfont] at (0,-2.56) {How many bears walk down the driveway?\\[3pt]\textcolor{ink}{\ding{51}\;Five: one adult and four cubs.}};
\end{scope}

\begin{scope}[shift={(0.00,-8.30)}]
\node[anchor=north west, inner sep=0, font=\sffamily\fontsize{9}{10.5}\selectfont\bfseries, text=catR] at (0,0) {(e) Reasoning: R1 Causality};
\node[anchor=north west, inner sep=0] (fe0) at (0.00,-0.43) {\includegraphics[width=2.5cm,height=1.40625cm]{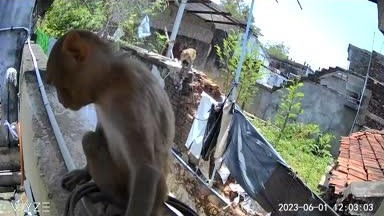}};
\node[anchor=north west, fill=black!70, text=white, font=\sffamily\fontsize{7}{8}\selectfont, inner sep=1.4pt] at (fe0.north west) {0:04};
\node[anchor=north, text width=2.5cm, align=center, inner sep=0, font=\sffamily\fontsize{7.3}{8.4}\selectfont\itshape, text=ink] at ($(fe0.south)+(0,-0.09)$) {Monkey sits by the camera};
\node[anchor=north west, inner sep=0] (fe1) at (2.60,-0.43) {\includegraphics[width=2.5cm,height=1.40625cm]{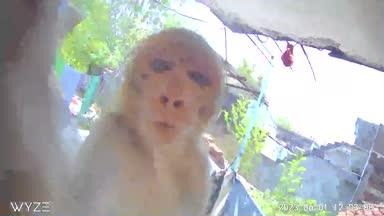}};
\node[anchor=north west, fill=black!70, text=white, font=\sffamily\fontsize{7}{8}\selectfont, inner sep=1.4pt] at (fe1.north west) {0:08};
\node[anchor=north, text width=2.5cm, align=center, inner sep=0, font=\sffamily\fontsize{7.3}{8.4}\selectfont\itshape, text=ink] at ($(fe1.south)+(0,-0.09)$) {Grabs the camera};
\node[anchor=north west, inner sep=0] (fe2) at (5.20,-0.43) {\includegraphics[width=2.5cm,height=1.40625cm]{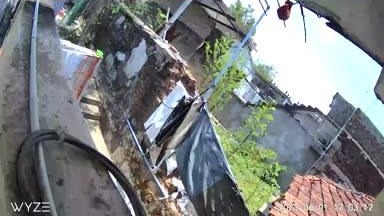}};
\node[anchor=north west, fill=black!70, text=white, font=\sffamily\fontsize{7}{8}\selectfont, inner sep=1.4pt] at (fe2.north west) {0:13};
\node[anchor=north, text width=2.5cm, align=center, inner sep=0, font=\sffamily\fontsize{7.3}{8.4}\selectfont\itshape, text=ink] at ($(fe2.south)+(0,-0.09)$) {View is left tilted};
\node[anchor=north west, draw=catR!35, fill=catR!4, rounded corners=2pt, text width=7.35cm, minimum height=1.05cm, inner sep=4.5pt, align=left, font=\sffamily\fontsize{8.5}{10}\selectfont] at (0,-2.56) {Why does the camera view tilt?\\[3pt]\textcolor{ink}{\ding{51}\;A monkey pushed the camera.}};
\end{scope}

\begin{scope}[shift={(8.15,-8.30)}]
\node[anchor=north west, inner sep=0, font=\sffamily\fontsize{9}{10.5}\selectfont\bfseries, text=catR] at (0,0) {(f) Reasoning: R2 Spatial};
\node[anchor=north west, inner sep=0] (ff0) at (0.00,-0.43) {\includegraphics[width=2.5cm,height=1.40625cm]{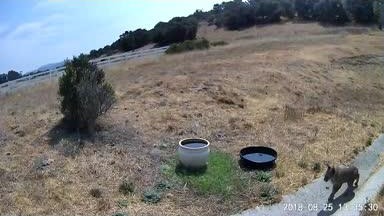}};
\node[anchor=north west, fill=black!70, text=white, font=\sffamily\fontsize{7}{8}\selectfont, inner sep=1.4pt] at (ff0.north west) {0:04};
\node[anchor=north, text width=2.5cm, align=center, inner sep=0, font=\sffamily\fontsize{7.3}{8.4}\selectfont\itshape, text=ink] at ($(ff0.south)+(0,-0.09)$) {Enters along the curb};
\node[anchor=north west, inner sep=0] (ff1) at (2.60,-0.43) {\includegraphics[width=2.5cm,height=1.40625cm]{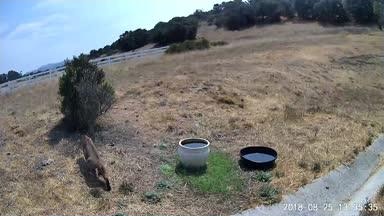}};
\node[anchor=north west, fill=black!70, text=white, font=\sffamily\fontsize{7}{8}\selectfont, inner sep=1.4pt] at (ff1.north west) {0:09};
\node[anchor=north, text width=2.5cm, align=center, inner sep=0, font=\sffamily\fontsize{7.3}{8.4}\selectfont\itshape, text=ink] at ($(ff1.south)+(0,-0.09)$) {Heads for the bush};
\node[anchor=north west, inner sep=0] (ff2) at (5.20,-0.43) {\includegraphics[width=2.5cm,height=1.40625cm]{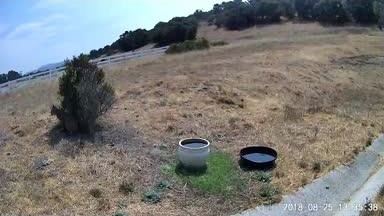}};
\node[anchor=north west, fill=black!70, text=white, font=\sffamily\fontsize{7}{8}\selectfont, inner sep=1.4pt] at (ff2.north west) {0:12};
\node[anchor=north, text width=2.5cm, align=center, inner sep=0, font=\sffamily\fontsize{7.3}{8.4}\selectfont\itshape, text=ink] at ($(ff2.south)+(0,-0.09)$) {Lies down under the bush};
\node[anchor=north west, draw=catR!35, fill=catR!4, rounded corners=2pt, text width=7.35cm, minimum height=1.05cm, inner sep=4.5pt, align=left, font=\sffamily\fontsize{8.5}{10}\selectfont] at (0,-2.56) {Where does the bobcat hide at the end of the clip?\\[3pt]\textcolor{ink}{\ding{51}\;Under the bush, in its shade.}};
\end{scope}

\begin{scope}[shift={(0.00,-12.45)}]
\node[anchor=north west, inner sep=0, font=\sffamily\fontsize{9}{10.5}\selectfont\bfseries, text=catR] at (0,0) {(g) Reasoning: R3 Temporal};
\node[anchor=north west, inner sep=0] (fg0) at (0.00,-0.43) {\includegraphics[width=2.5cm,height=1.40625cm]{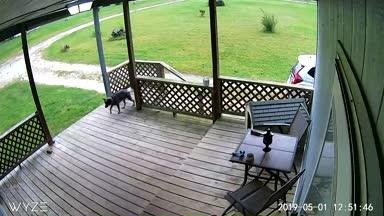}};
\node[anchor=north west, fill=black!70, text=white, font=\sffamily\fontsize{7}{8}\selectfont, inner sep=1.4pt] at (fg0.north west) {0:01};
\node[anchor=north, text width=2.5cm, align=center, inner sep=0, font=\sffamily\fontsize{7.3}{8.4}\selectfont\itshape, text=ink] at ($(fg0.south)+(0,-0.09)$) {Dog climbs onto the porch};
\node[anchor=north west, inner sep=0] (fg1) at (2.60,-0.43) {\includegraphics[width=2.5cm,height=1.40625cm]{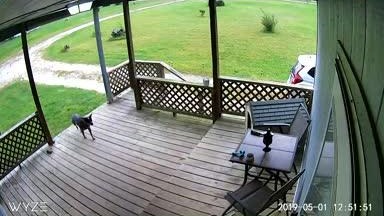}};
\node[anchor=north west, fill=black!70, text=white, font=\sffamily\fontsize{7}{8}\selectfont, inner sep=1.4pt] at (fg1.north west) {0:06};
\node[anchor=north, text width=2.5cm, align=center, inner sep=0, font=\sffamily\fontsize{7.3}{8.4}\selectfont\itshape, text=ink] at ($(fg1.south)+(0,-0.09)$) {Sniffs around the deck};
\node[anchor=north west, inner sep=0] (fg2) at (5.20,-0.43) {\includegraphics[width=2.5cm,height=1.40625cm]{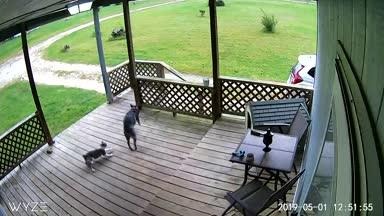}};
\node[anchor=north west, fill=black!70, text=white, font=\sffamily\fontsize{7}{8}\selectfont, inner sep=1.4pt] at (fg2.north west) {0:09};
\node[anchor=north, text width=2.5cm, align=center, inner sep=0, font=\sffamily\fontsize{7.3}{8.4}\selectfont\itshape, text=ink] at ($(fg2.south)+(0,-0.09)$) {Cat charges at the dog};
\node[anchor=north west, draw=catR!35, fill=catR!4, rounded corners=2pt, text width=7.35cm, minimum height=1.05cm, inner sep=4.5pt, align=left, font=\sffamily\fontsize{8.5}{10}\selectfont] at (0,-2.56) {What does the dog do just before the cat charges?\\[3pt]\textcolor{ink}{\ding{51}\;Sniffs around the porch.}};
\end{scope}

\begin{scope}[shift={(8.15,-12.45)}]
\node[anchor=north west, inner sep=0, font=\sffamily\fontsize{9}{10.5}\selectfont\bfseries, text=catS] at (0,0) {(h) Insight: I1 Intent};
\node[anchor=north west, inner sep=0] (fh0) at (0.00,-0.43) {\includegraphics[width=2.5cm,height=1.40625cm]{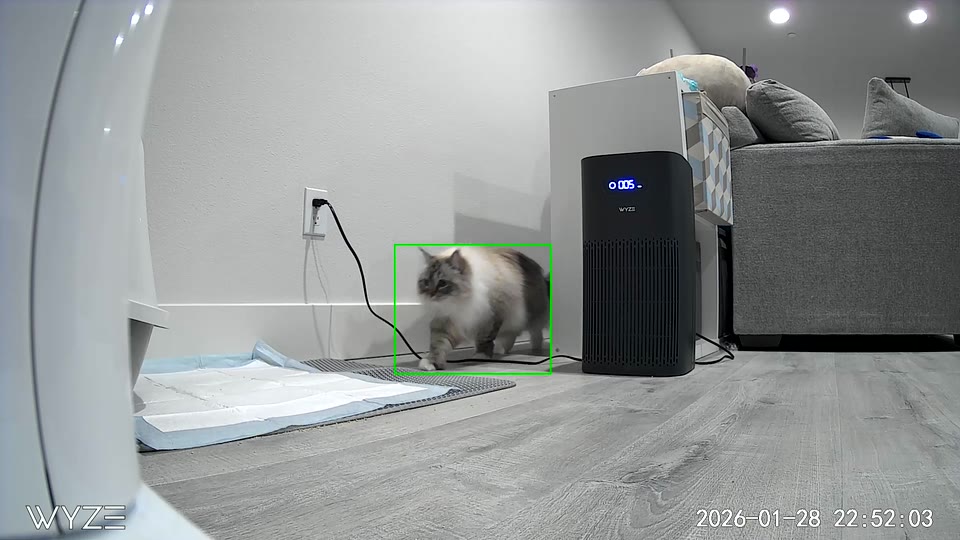}};
\node[anchor=north west, fill=black!70, text=white, font=\sffamily\fontsize{7}{8}\selectfont, inner sep=1.4pt] at (fh0.north west) {0:01};
\node[anchor=north, text width=2.5cm, align=center, inner sep=0, font=\sffamily\fontsize{7.3}{8.4}\selectfont\itshape, text=ink] at ($(fh0.south)+(0,-0.09)$) {Cat walks towards the pad};
\node[anchor=north west, inner sep=0] (fh1) at (2.60,-0.43) {\includegraphics[width=2.5cm,height=1.40625cm]{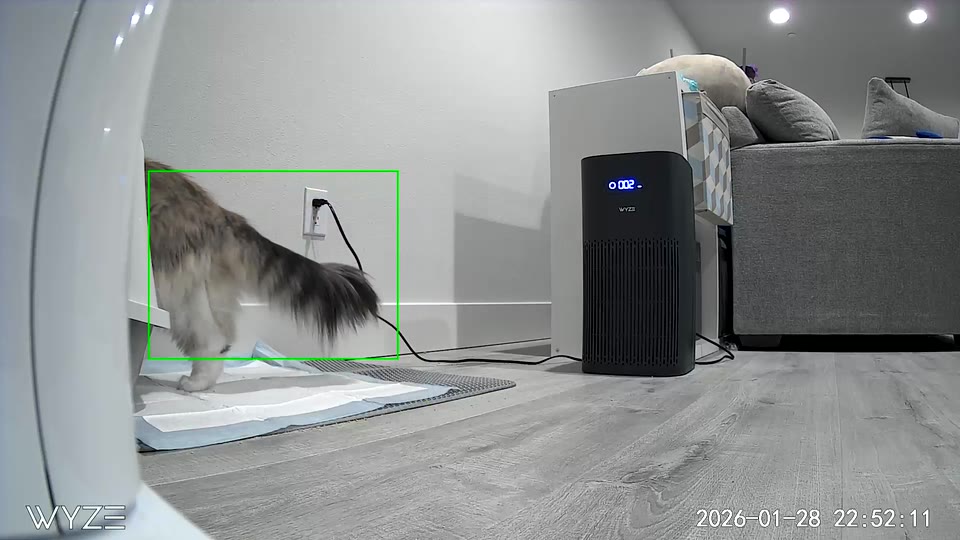}};
\node[anchor=north west, fill=black!70, text=white, font=\sffamily\fontsize{7}{8}\selectfont, inner sep=1.4pt] at (fh1.north west) {0:08};
\node[anchor=north, text width=2.5cm, align=center, inner sep=0, font=\sffamily\fontsize{7.3}{8.4}\selectfont\itshape, text=ink] at ($(fh1.south)+(0,-0.09)$) {Steps onto the pad};
\node[anchor=north west, inner sep=0] (fh2) at (5.20,-0.43) {\includegraphics[width=2.5cm,height=1.40625cm]{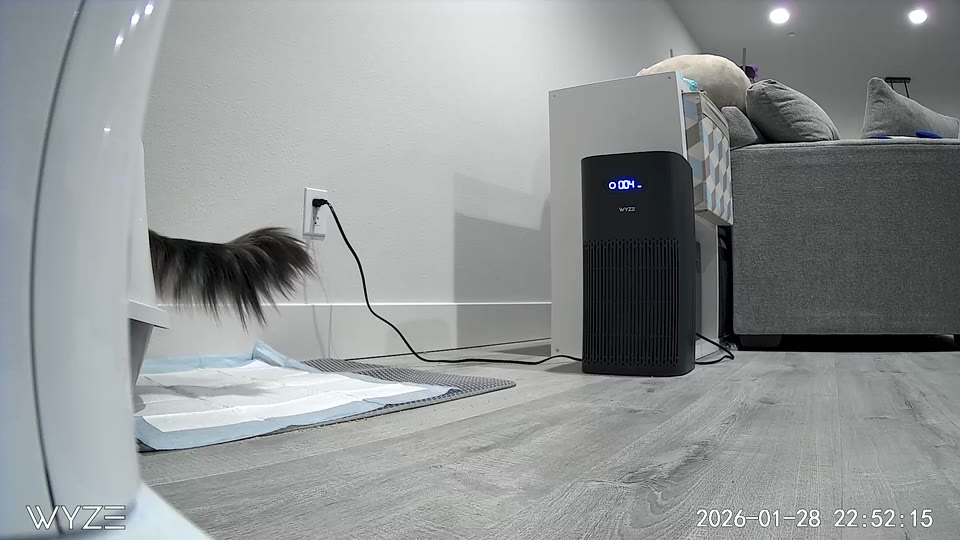}};
\node[anchor=north west, fill=black!70, text=white, font=\sffamily\fontsize{7}{8}\selectfont, inner sep=1.4pt] at (fh2.north west) {0:12};
\node[anchor=north, text width=2.5cm, align=center, inner sep=0, font=\sffamily\fontsize{7.3}{8.4}\selectfont\itshape, text=ink] at ($(fh2.south)+(0,-0.09)$) {Exits the frame};
\node[anchor=north west, draw=catS!35, fill=catS!4, rounded corners=2pt, text width=7.35cm, minimum height=1.05cm, inner sep=4.5pt, align=left, font=\sffamily\fontsize{8.5}{10}\selectfont] at (0,-2.56) {What is the intent of the animal in this video?\\[3pt]\textcolor{ink}{\ding{51}\;To use the litter box.}};
\end{scope}

\begin{scope}[shift={(0.00,-16.60)}]
\node[anchor=north west, inner sep=0, font=\sffamily\fontsize{9}{10.5}\selectfont\bfseries, text=catS] at (0,0) {(i) Insight: I2 Evidence};
\node[anchor=north west, inner sep=0] (fi0) at (0.00,-0.43) {\includegraphics[width=2.5cm,height=1.40625cm]{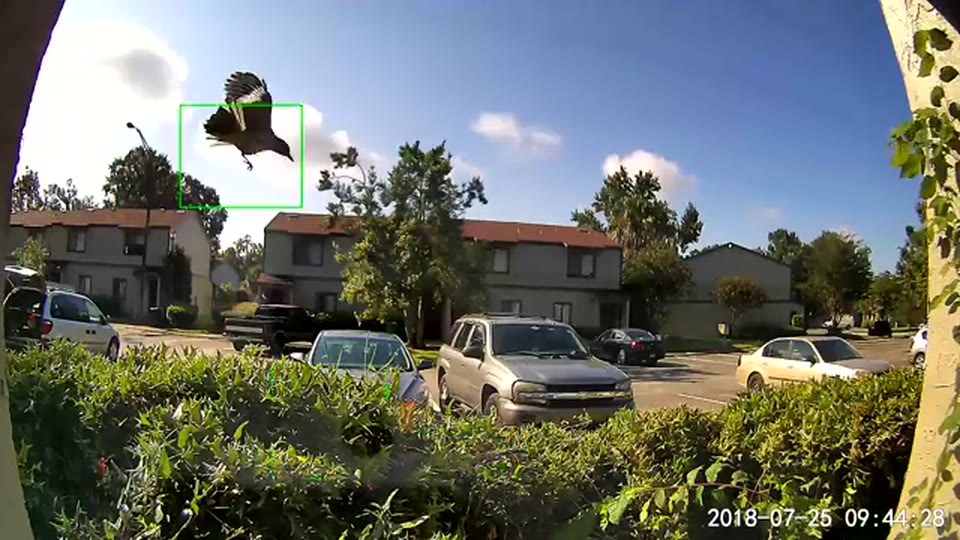}};
\node[anchor=north west, fill=black!70, text=white, font=\sffamily\fontsize{7}{8}\selectfont, inner sep=1.4pt] at (fi0.north west) {0:02};
\node[anchor=north, text width=2.5cm, align=center, inner sep=0, font=\sffamily\fontsize{7.3}{8.4}\selectfont\itshape, text=ink] at ($(fi0.south)+(0,-0.09)$) {Bird dives towards the hedge};
\node[anchor=north west, inner sep=0] (fi1) at (2.60,-0.43) {\includegraphics[width=2.5cm,height=1.40625cm]{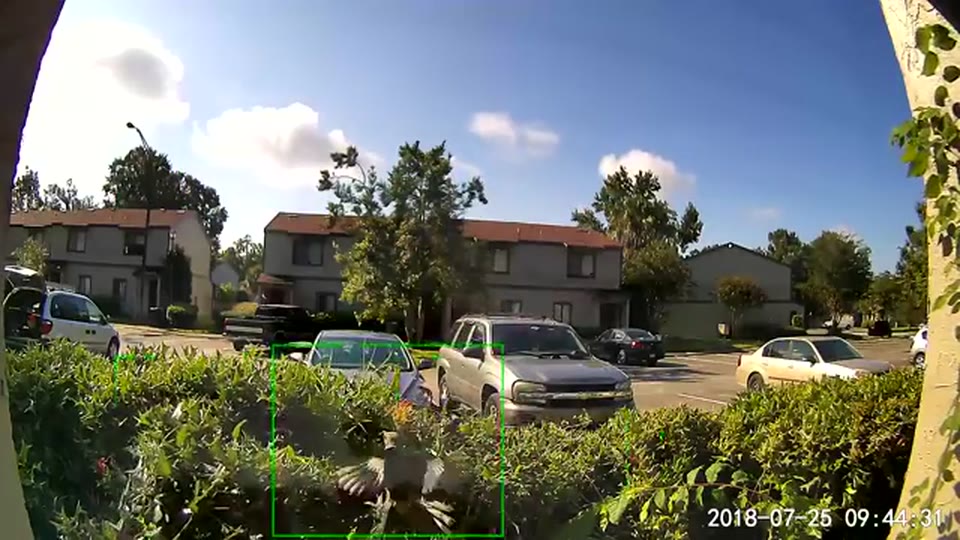}};
\node[anchor=north west, fill=black!70, text=white, font=\sffamily\fontsize{7}{8}\selectfont, inner sep=1.4pt] at (fi1.north west) {0:05};
\node[anchor=north, text width=2.5cm, align=center, inner sep=0, font=\sffamily\fontsize{7.3}{8.4}\selectfont\itshape, text=ink] at ($(fi1.south)+(0,-0.09)$) {Strikes into the hedge};
\node[anchor=north west, inner sep=0] (fi2) at (5.20,-0.43) {\includegraphics[width=2.5cm,height=1.40625cm]{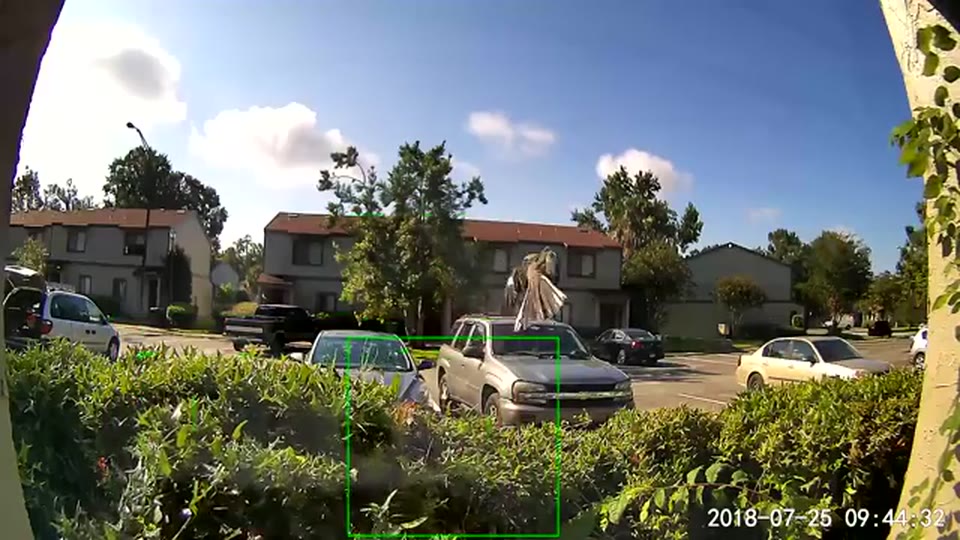}};
\node[anchor=north west, fill=black!70, text=white, font=\sffamily\fontsize{7}{8}\selectfont, inner sep=1.4pt] at (fi2.north west) {0:06};
\node[anchor=north, text width=2.5cm, align=center, inner sep=0, font=\sffamily\fontsize{7.3}{8.4}\selectfont\itshape, text=ink] at ($(fi2.south)+(0,-0.09)$) {Rises to strike again};
\node[anchor=north west, draw=catS!35, fill=catS!4, rounded corners=2pt, text width=7.35cm, minimum height=1.05cm, inner sep=4.5pt, align=left, font=\sffamily\fontsize{8.5}{10}\selectfont] at (0,-2.56) {What is the evidence of the bird's intent?\\[3pt]\textcolor{ink}{\ding{51}\;It keeps diving at the snake in the hedge to attack it.}};
\end{scope}

\begin{scope}[shift={(8.15,-16.60)}]
\node[anchor=north west, inner sep=0, font=\sffamily\fontsize{9}{10.5}\selectfont\bfseries, text=catS] at (0,0) {(j) Insight: I3 Recommendation};
\node[anchor=north west, inner sep=0] (fj0) at (0.00,-0.43) {\includegraphics[width=2.5cm,height=1.40625cm]{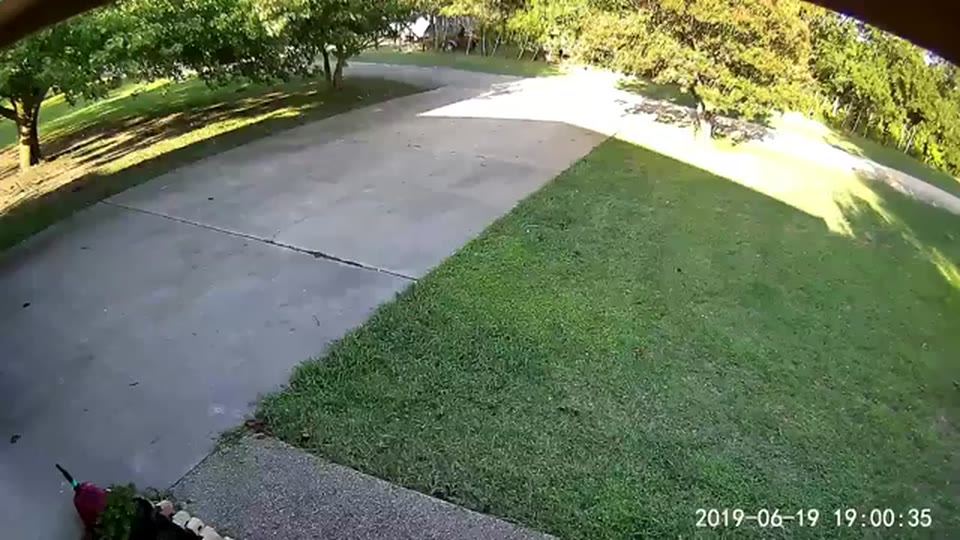}};
\node[anchor=north west, fill=black!70, text=white, font=\sffamily\fontsize{7}{8}\selectfont, inner sep=1.4pt] at (fj0.north west) {0:01};
\node[anchor=north, text width=2.5cm, align=center, inner sep=0, font=\sffamily\fontsize{7.3}{8.4}\selectfont\itshape, text=ink] at ($(fj0.south)+(0,-0.09)$) {Driveway and lawn are clear};
\node[anchor=north west, inner sep=0] (fj1) at (2.60,-0.43) {\includegraphics[width=2.5cm,height=1.40625cm]{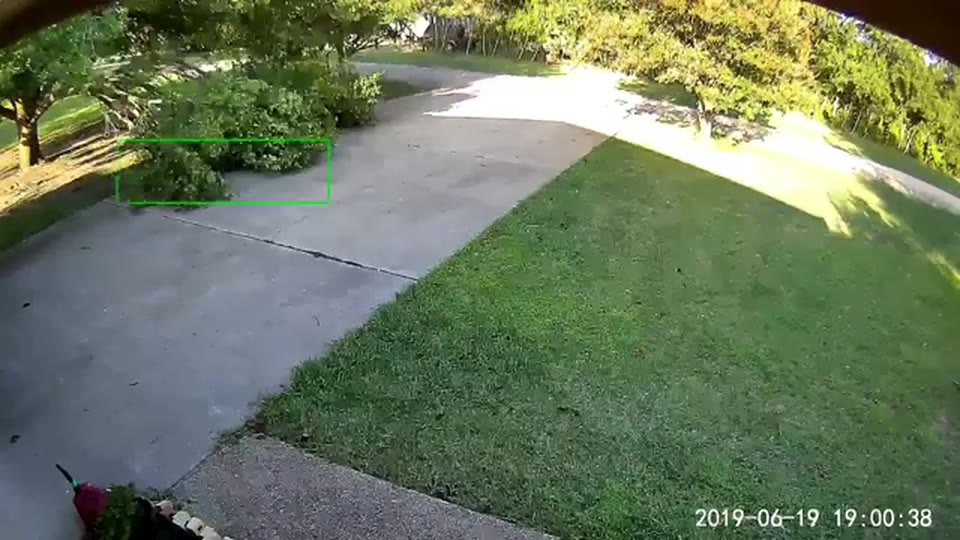}};
\node[anchor=north west, fill=black!70, text=white, font=\sffamily\fontsize{7}{8}\selectfont, inner sep=1.4pt] at (fj1.north west) {0:04};
\node[anchor=north, text width=2.5cm, align=center, inner sep=0, font=\sffamily\fontsize{7.3}{8.4}\selectfont\itshape, text=ink] at ($(fj1.south)+(0,-0.09)$) {A large branch falls};
\node[anchor=north west, inner sep=0] (fj2) at (5.20,-0.43) {\includegraphics[width=2.5cm,height=1.40625cm]{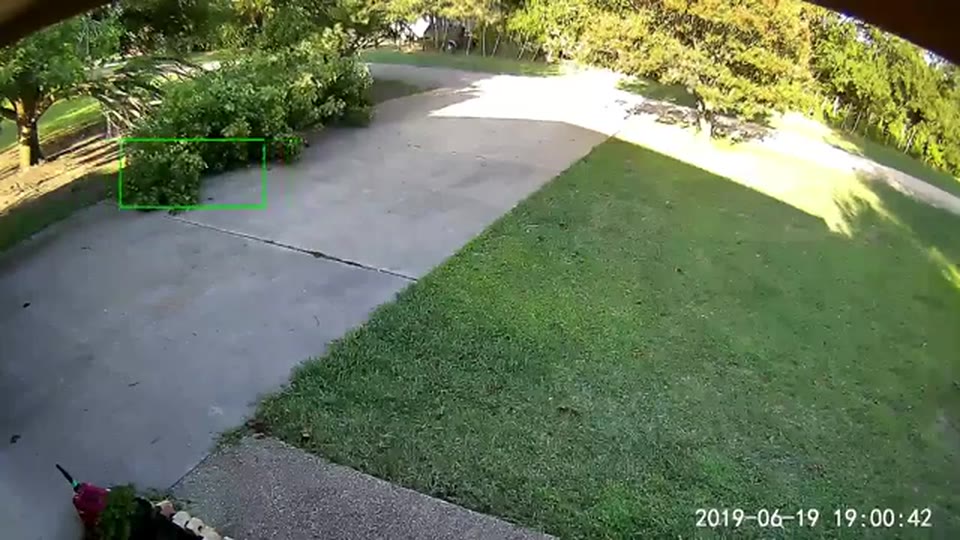}};
\node[anchor=north west, fill=black!70, text=white, font=\sffamily\fontsize{7}{8}\selectfont, inner sep=1.4pt] at (fj2.north west) {0:07};
\node[anchor=north, text width=2.5cm, align=center, inner sep=0, font=\sffamily\fontsize{7.3}{8.4}\selectfont\itshape, text=ink] at ($(fj2.south)+(0,-0.09)$) {Branch lies across the driveway};
\node[anchor=north west, draw=catS!35, fill=catS!4, rounded corners=2pt, text width=7.35cm, minimum height=1.05cm, inner sep=4.5pt, align=left, font=\sffamily\fontsize{8.5}{10}\selectfont] at (0,-2.56) {What should the homeowner do given this event?\\[3pt]\textcolor{ink}{\ding{51}\;Arrange to remove the fallen branch.}};
\end{scope}
\end{tikzpicture}}
\caption{\textbf{Example clips and questions for the 10 tasks.} Panels (a)--(d) show Perception, (e)--(g) Reasoning, and (h)--(j) Insight. Each example includes three key frames with timestamps and a question with its correct answer. Answer options are omitted for readability; each benchmark question has four options.}
\label{fig:example}
\end{figure}
 
\subsection{Data collection}
\label{sec:collection}
\bench{} clips are sourced from the Wyze user forum, a public community where homeowners share
their smart-home camera footage. Clips were selected to cover challenging
smart-home camera conditions, including daytime and night-time footage in colour and infrared,
predominantly outdoor views, recordings that begin or end mid-action, and events involving people,
vehicles, animals and natural phenomena. Detailed dataset statistics are given in
\cref{tab:stats} (\cref{sec:stats}).

\paragraph{Licensing and privacy.} The clips and associated question--answer pairs are licensed under the Creative Commons Attribution-NonCommercial 4.0 International (CC BY-NC 4.0) licence.

\subsection{Question construction and annotation}
\label{sec:annotation}

We construct the question answer set by combining automated question generation with human
verification and refinement. The process uses the event descriptions provided by forum users as contextual guidance, while requiring every question and answer to be
grounded in visual evidence. The resulting benchmark contains 2,254 questions
across 1,118 clips, with an average of 2.02 questions per clip. Question
development proceeds in four stages.

Each forum post pairs its clip with the homeowner's description of the event, ranging from
interpretations such as ``a thief on my porch'' or ``pet nuisance on my lawn'' to neutral
observations such as ``bright sunny day.'' These descriptions provide useful context for question
generation but are not reliable ground truth: they may describe events outside the frame, infer
subject intent that cannot be established visually, or rely on contextual knowledge unavailable
from the clip, such as a person's identity. Post text therefore
guides question generation, but all questions and answers must be grounded in visual evidence.
Post text is used only during benchmark construction and is never shown to evaluated models
(\cref{sec:protocol}).

\begin{enumerate}[leftmargin=1.6em, label=\textbf{\arabic*.}]
  \item \textbf{Grounded drafting.} Each clip and its post text are inputs to the Gemini 2.5 Pro, Gemini 3.8 Flash and GPT-6 Astra application programming interfaces (APIs). The models first generate a description of the clip and then
        draft multiple-choice questions, each with four answer options, for the tasks relevant to the clip. The post text serves as a guide
        to potentially relevant events; but the models are instructed to ground every question in what is
        visible, not in what the post claims.
  \item \textbf{Question refinement.} Claude Sonnet 4.5 rewrites each
        question and its options to eliminate shortcuts. Distractors are designed as \emph{near
        misses}: plausible given the scene, but differing from the correct answer in a single
        detail that must be checked against the clip. Options are matched in length to prevent answer length from revealing
        the correct choice. Each question is then
        put to Gemini 2.5 Pro, GPT-6.1 Sol and Gemma 4 31B \emph{without the video}. If all three
        models answer correctly from the text alone, the question is flagged as text-answerable and refined
        using their text-only reasoning as feedback. Their explanations help identify wording or options that reveal the
        answer without visual inspection. If all three models still answer
        correctly without the video after a second refinement, the question is
        discarded. Each question has four options (A--D), with the correct
        answer position approximately balanced across the four choices.
\item \textbf{Human fact verification.} A nine-person annotation team
    verifies the factual claims in the model-generated clip descriptions,
    questions, and answer options against the video. The annotation team provides corrections and feedback to ensure they are factually accurate and
grounded in visual evidence.
  \item \textbf{Revision from human feedback.} Questions and answers are revised based on annotator feedback: ambiguous questions are reworded, incorrect answers fixed. The team then reviews 25\% of the final generated
        questions.
\end{enumerate}

Together, these stages combine automated question generation with visual
fact verification and human review. The text-only screening filters questions
that can be answered without access to the video, while balancing the correct
answer position guards against answer-position bias. The best constant-answer baseline achieves 26.9\%, close to the
25\% chance level (\cref{sec:protocol}).

\subsection{Statistics}
\label{sec:stats}

\Cref{tab:stats} summarises the benchmark, overall and by capability level: 2,254 questions in 1,118 clips.
The clips total 5.97 hours of footage. They are short, with a mean duration of 19.2\,s; the
longest is 275\,s. Although 86\% of source clips are 1080p or higher, evaluation uses resized frames
(\cref{sec:protocol}). Source resolution alone therefore does not establish whether image quality
limits performance; viewpoint, lighting, partial capture and input resizing may all contribute.
A clip can carry questions from multiple levels, so the per-level clip counts and hours in
\cref{tab:stats} sum to more than the overall totals.

\paragraph{Lighting and setting.} Day and night footage are nearly evenly represented: accounting for 50\% and 46\% of questions, respectively; dusk
accounts for the remainder. The night share includes infrared footage,
which represents 22\% of all questions. Night footage accounts for majority of questions in
Recognition (P2) and Count (P4), but only about a third in Attribute (P1); Insight sits in between, with roughly two in five questions at night. Indoor
cameras are rare, contributing fewer than one question in ten. 

\paragraph{Subjects.} Animals are the most frequently present subject, appearing in more than half the
questions, followed by vehicles (over two in five) and people (about a quarter). The subject mix varies by level: animals are the main subject of about three in five Perception and
Reasoning questions, whereas Insight centres on people, who are the main subject of nearly half its
questions, in line with Insight's focus on deliveries, visitors and
possible thefts.

\begin{table}[t]
\centering
\caption{\textbf{\bench{} at a glance.} Questions, clips, footage, scene conditions and subjects
present, per capability level and in total. Scene and Subjects Present columns are shares of questions
(\cref{sec:stats}).}
\label{tab:stats}
\resizebox{\textwidth}{!}{%
\begin{tabular}{@{}l c r r r r r r r r r r r@{}}
\toprule
& & & & & & & \multicolumn{3}{c}{\textbf{Scene}} & \multicolumn{3}{c}{\textbf{Subjects Present}} \\
\cmidrule(lr){8-10}\cmidrule(l){11-13}
\textbf{Capability Level} & \textbf{Tasks} & \textbf{Questions} & \textbf{Share} & \textbf{Clips} & \textbf{Hours} & \textbf{Questions / clip} & \textbf{Day} & \textbf{Night (IR)} & \textbf{Indoor} & \textbf{Person} & \textbf{Vehicle} & \textbf{Animal} \\
\midrule
1\;\;Perception & 4 & 989 & 43.9\% & 703 & 3.57 & 1.4 & 46\% & 50\% (27\%) & 9\% & 19\% & 40\% & 62\% \\
2\;\;Reasoning  & 3 & 721 & 32.0\% & 543 & 2.75 & 1.3 & 50\% & 46\% (23\%) & 8\% & 14\% & 40\% & 63\% \\
3\;\;Insight    & 3 & 544 & 24.1\% & 240 & 1.51 & 2.3 & 57\% & 40\% (13\%) & 8\% & 57\% & 51\% & 35\% \\
\midrule
\textbf{Total} & \textbf{10} & \textbf{2,254} & 100\% & \textbf{1,118} & \textbf{5.97} & \textbf{2.0} & \textbf{50\%} & \textbf{46\% (22\%)} & \textbf{8\%} & \textbf{26\%} & \textbf{43\%} & \textbf{56\%} \\
\bottomrule
\end{tabular}}
\end{table}

\section{Evaluation Protocol}
\label{sec:protocol}
\paragraph{Input.} Each model receives the clip and a single question with its four answer options.
All models, whether accessed through an API or served locally with SGLang or vLLM, follow the
same evaluation protocol. Frames are sampled at 1 frame per second (fps), up to a maximum of 64 frames per clip, and each frame
is resized to a budget of 200{,}000 pixels (approximately $596\times335$ for a 16:9 clip). The 64-frame cap affects clips longer than 64s; for these clips, frames are sampled uniformly across the entire clip. \Cref{tab:full} lists the input and settings of every model. Every question is
answerable from the visual content, and no audio, subtitles, transcripts or metadata are provided.

\paragraph{Decoding and reasoning.} We use the minimum reasoning setting available for each
model. For API-based models, reasoning is disabled or set to the lowest available level
(e.g., \emph{minimal} or \emph{low}), and temperature is set to zero or the lowest value
permitted by the API. For locally served models with a switchable reasoning mode, explicit
reasoning is disabled by passing \texttt{enable\_thinking=False} to the chat template. These settings constrain explicit reasoning during evaluation;
the results therefore reflect performance under the specified decoding and reasoning settings.

\paragraph{Metrics.} We report accuracy for each of the 10 tasks. A \emph{capability-level score} is the
accuracy over all questions at that level, and the \emph{overall score} is the accuracy over all
2,254 questions; both weight each task by its number of questions (\cref{tab:pertask}).
Each question is scored independently.

\paragraph{Controls.} As a check on answer-position bias, we score a constant-answer baseline that
picks the same option letter for every question. With four options, chance is 25\%; the best
such letter scores 26.9\% overall, so correct answers are close to evenly spread across positions
and a model cannot do well by favouring one letter.

\section{Experiments}
\label{sec:experiments}

\subsection{Models}
We evaluate 32 models spanning the GPT, Gemini, Grok, Gemma, Qwen, MiMo and Meta Muse
families; \cref{tab:full} (\cref{app:results}) lists each model's input configuration and evaluation settings. All 32
have complete results across all three capability levels.

\begin{table}[!t]
\centering
\renewcommand{\arraystretch}{1.06}
\caption{\textbf{Per-task accuracy (\%)} of selected closed-source and open-weight models on \bench{},
each group sorted by overall accuracy. The second header row gives the number of questions per task. Within each group, the best score per column is in
\textbf{bold} and the second best \underline{underlined}. Results for all 32 models are in \cref{tab:full}.}
\label{tab:pertask}
\resizebox{\textwidth}{!}{%
\begin{tabular}{@{}l *{4}{c} | *{3}{c} | *{3}{c}@{}}
\toprule
& \multicolumn{4}{c|}{\textbf{Perception}} & \multicolumn{3}{c|}{\textbf{Reasoning}} & \multicolumn{3}{c}{\textbf{Insight}} \\
\textbf{Model} & P1 & P2 & P3 & P4 & R1 & R2 & R3 & I1 & I2 & I3 \\
\textcolor{muted}{\# questions} & \textcolor{muted}{249} & \textcolor{muted}{246} & \textcolor{muted}{249} & \textcolor{muted}{245} & \textcolor{muted}{248} & \textcolor{muted}{237} & \textcolor{muted}{236} & \textcolor{muted}{187} & \textcolor{muted}{186} & \textcolor{muted}{171} \\
\midrule
\multicolumn{11}{@{}l}{\textit{Closed-source models}}\\
Gemini 4 Argon & \textbf{89.2} & \textbf{78.5} & \textbf{92.4} & \textbf{66.1} & \textbf{86.3} & \textbf{71.7} & \textbf{65.7} & \textbf{87.2} & \textbf{90.9} & \textbf{89.5} \\
GPT-6 Astra & 83.9 & \underline{72.8} & \underline{87.2} & \underline{61.6} & \underline{79.0} & \underline{70.5} & \underline{56.8} & \underline{84.5} & \underline{89.8} & \underline{86.5} \\
Gemini 3.8 Flash & 82.3 & 65.5 & 84.7 & 50.6 & 75.8 & 61.6 & 51.3 & 80.2 & 79.6 & 77.2 \\
Muse Wonder 1.0 & \underline{84.3} & 61.4 & 81.9 & 51.4 & 73.0 & 63.3 & 49.6 & 68.4 & 78.5 & 74.3 \\
Gemini 3.1 Pro & 79.9 & 58.5 & 75.1 & 46.1 & 73.4 & 59.5 & 45.8 & 64.7 & 68.8 & 66.7 \\
Gemini 2.5 Pro & 80.3 & 59.4 & 73.5 & 40.4 & 66.9 & 46.0 & 43.2 & 52.9 & 59.7 & 61.4 \\
Grok 4.7 & 77.1 & 45.9 & 74.7 & 40.4 & 64.1 & 50.6 & 45.3 & 50.8 & 71.0 & 50.3 \\
Gemini 3.5 Flash-Lite & 77.1 & 55.3 & 65.9 & 42.9 & 67.7 & 43.0 & 34.8 & 51.3 & 54.3 & 56.7 \\
GPT-4o & 74.3 & 52.8 & 66.7 & 33.1 & 65.7 & 37.1 & 38.1 & 51.3 & 60.8 & 55.0 \\
GPT-5 mini & 79.9 & 54.5 & 67.9 & 33.9 & 60.1 & 45.6 & 41.1 & 45.5 & 38.2 & 42.7 \\
\midrule
\multicolumn{11}{@{}l}{\textit{Open-weight models}}\\
Muse-Glimmer-30B & \textbf{81.9} & \textbf{55.3} & \underline{77.1} & \textbf{46.5} & 63.3 & \textbf{50.2} & \textbf{43.2} & \underline{48.1} & \underline{52.2} & \textbf{55.0} \\
Gemma 4 31B-it & \underline{78.7} & \underline{45.9} & \textbf{79.9} & \underline{43.3} & \underline{63.7} & \underline{47.7} & \underline{42.4} & 46.0 & 44.1 & 46.1 \\
Qwen3.5-397B-A17B & 76.3 & 44.7 & 71.9 & 32.7 & 62.1 & 38.4 & 40.3 & 45.5 & \textbf{52.7} & \underline{52.6} \\
Qwen3.8-Flash-Next & 73.5 & 43.1 & 71.5 & 35.1 & \textbf{65.7} & 45.1 & 38.1 & \textbf{50.8} & 45.7 & 42.7 \\
Qwen3.5-27B & 70.3 & 40.2 & 68.7 & 34.3 & 62.9 & 38.8 & 34.3 & 44.4 & 48.9 & 44.4 \\
MiMo-V2.6-Pro-MOPD & 66.7 & 32.5 & 67.5 & 35.1 & 55.6 & 35.4 & 33.5 & \underline{48.1} & 47.8 & 41.5 \\
MiMo-V2.6-Flash-MOPD & 66.3 & 34.6 & 55.4 & 32.7 & 55.2 & 39.7 & 26.3 & 41.7 & 39.2 & 35.7 \\
\bottomrule
\end{tabular}}
\end{table}

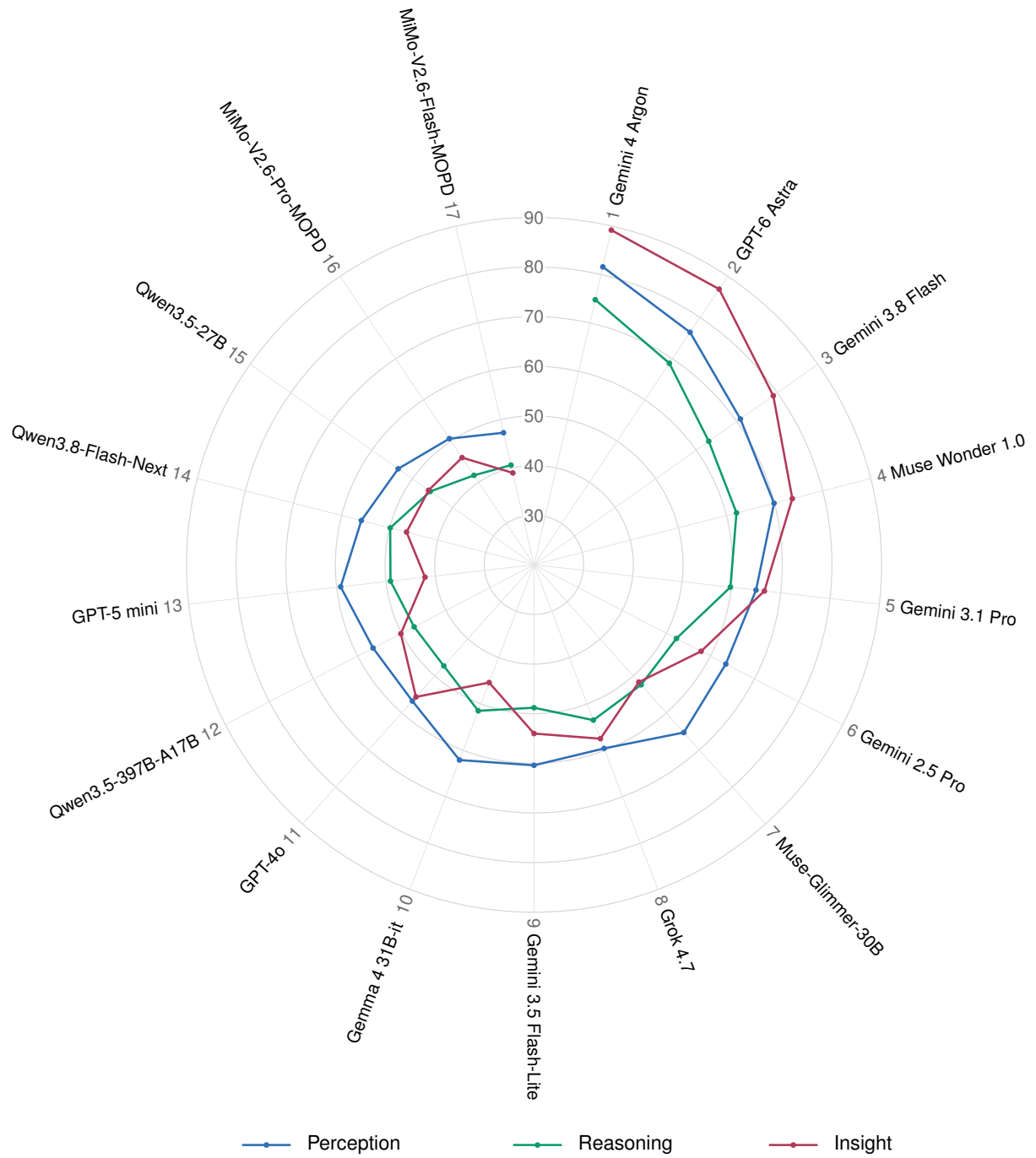
\begin{figure}[p]
\centering
\resizebox{\textwidth}{!}{%
\begin{tikzpicture}
\draw[rule, line width=0.45pt] (0,0) circle (0.714);
\node[font=\sffamily\scriptsize, text=muted, fill=white, inner sep=0.8pt] at (0,0.714) {30};
\draw[rule, line width=0.45pt] (0,0) circle (1.429);
\node[font=\sffamily\scriptsize, text=muted, fill=white, inner sep=0.8pt] at (0,1.429) {40};
\draw[rule, line width=0.45pt] (0,0) circle (2.143);
\node[font=\sffamily\scriptsize, text=muted, fill=white, inner sep=0.8pt] at (0,2.143) {50};
\draw[rule, line width=0.45pt] (0,0) circle (2.857);
\node[font=\sffamily\scriptsize, text=muted, fill=white, inner sep=0.8pt] at (0,2.857) {60};
\draw[rule, line width=0.45pt] (0,0) circle (3.571);
\node[font=\sffamily\scriptsize, text=muted, fill=white, inner sep=0.8pt] at (0,3.571) {70};
\draw[rule, line width=0.45pt] (0,0) circle (4.286);
\node[font=\sffamily\scriptsize, text=muted, fill=white, inner sep=0.8pt] at (0,4.286) {80};
\draw[rule, line width=0.45pt] (0,0) circle (5.000);
\node[font=\sffamily\scriptsize, text=muted, fill=white, inner sep=0.8pt] at (0,5.000) {90};
\draw[rule!70, line width=0.3pt] (0,0) -- (1.125,4.872);
\node[anchor=west, rotate=77.00, font=\sffamily\scriptsize, inner sep=0pt] at (1.144,4.955) {\textcolor{muted}{1}\;Gemini 4 Argon};
\draw[rule!70, line width=0.3pt] (0,0) -- (2.787,4.151);
\node[anchor=west, rotate=56.12, font=\sffamily\scriptsize, inner sep=0pt] at (2.835,4.222) {\textcolor{muted}{2}\;GPT-6 Astra};
\draw[rule!70, line width=0.3pt] (0,0) -- (4.083,2.886);
\node[anchor=west, rotate=35.25, font=\sffamily\scriptsize, inner sep=0pt] at (4.153,2.935) {\textcolor{muted}{3}\;Gemini 3.8 Flash};
\draw[rule!70, line width=0.3pt] (0,0) -- (4.843,1.241);
\node[anchor=west, rotate=14.38, font=\sffamily\scriptsize, inner sep=0pt] at (4.926,1.263) {\textcolor{muted}{4}\;Muse Wonder 1.0};
\draw[rule!70, line width=0.3pt] (0,0) -- (4.968,-0.566);
\node[anchor=west, rotate=-6.50, font=\sffamily\scriptsize, inner sep=0pt] at (5.053,-0.576) {\textcolor{muted}{5}\;Gemini 3.1 Pro};
\draw[rule!70, line width=0.3pt] (0,0) -- (4.440,-2.299);
\node[anchor=west, rotate=-27.38, font=\sffamily\scriptsize, inner sep=0pt] at (4.516,-2.338) {\textcolor{muted}{6}\;Gemini 2.5 Pro};
\draw[rule!70, line width=0.3pt] (0,0) -- (3.329,-3.730);
\node[anchor=west, rotate=-48.25, font=\sffamily\scriptsize, inner sep=0pt] at (3.386,-3.794) {\textcolor{muted}{7}\;Muse-Glimmer-30B};
\draw[rule!70, line width=0.3pt] (0,0) -- (1.782,-4.672);
\node[anchor=west, rotate=-69.12, font=\sffamily\scriptsize, inner sep=0pt] at (1.812,-4.752) {\textcolor{muted}{8}\;Grok 4.7};
\draw[rule!70, line width=0.3pt] (0,0) -- (0.000,-5.000);
\node[anchor=west, rotate=-90.00, font=\sffamily\scriptsize, inner sep=0pt] at (0.000,-5.086) {\textcolor{muted}{9}\;Gemini 3.5 Flash-Lite};
\draw[rule!70, line width=0.3pt] (0,0) -- (-1.782,-4.672);
\node[anchor=east, rotate=69.12, font=\sffamily\scriptsize, inner sep=0pt] at (-1.812,-4.752) {Gemma 4 31B-it \;\textcolor{muted}{10}};
\draw[rule!70, line width=0.3pt] (0,0) -- (-3.329,-3.730);
\node[anchor=east, rotate=48.25, font=\sffamily\scriptsize, inner sep=0pt] at (-3.386,-3.794) {GPT-4o\;\textcolor{muted}{11}};
\draw[rule!70, line width=0.3pt] (0,0) -- (-4.440,-2.299);
\node[anchor=east, rotate=27.38, font=\sffamily\scriptsize, inner sep=0pt] at (-4.516,-2.338) {Qwen3.5-397B-A17B\;\textcolor{muted}{12}};
\draw[rule!70, line width=0.3pt] (0,0) -- (-4.968,-0.566);
\node[anchor=east, rotate=6.50, font=\sffamily\scriptsize, inner sep=0pt] at (-5.053,-0.576) {GPT-5 mini\;\textcolor{muted}{13}};
\draw[rule!70, line width=0.3pt] (0,0) -- (-4.843,1.241);
\node[anchor=east, rotate=-14.38, font=\sffamily\scriptsize, inner sep=0pt] at (-4.926,1.263) {Qwen3.8-Flash-Next\;\textcolor{muted}{14}};
\draw[rule!70, line width=0.3pt] (0,0) -- (-4.083,2.886);
\node[anchor=east, rotate=-35.25, font=\sffamily\scriptsize, inner sep=0pt] at (-4.153,2.935) {Qwen3.5-27B\;\textcolor{muted}{15}};
\draw[rule!70, line width=0.3pt] (0,0) -- (-2.787,4.151);
\node[anchor=east, rotate=-56.12, font=\sffamily\scriptsize, inner sep=0pt] at (-2.835,4.222) {MiMo-V2.6-Pro-MOPD\;\textcolor{muted}{16}};
\draw[rule!70, line width=0.3pt] (0,0) -- (-1.125,4.872);
\node[anchor=east, rotate=-77.00, font=\sffamily\scriptsize, inner sep=0pt] at (-1.144,4.955) {MiMo-V2.6-Flash-MOPD\;\textcolor{muted}{17}};
\draw[catP, line width=0.9pt, line join=round] (0.990,4.287) -- (2.245,3.345) -- (2.969,2.098) -- (3.453,0.885) -- (3.194,-0.364) -- (2.759,-1.429) -- (2.155,-2.414) -- (1.008,-2.643) -- (0.000,-2.886) -- (-1.072,-2.810) -- (-1.750,-1.961) -- (-2.315,-1.199) -- (-2.782,-0.317) -- (-2.484,0.637) -- (-1.954,1.381) -- (-1.218,1.815) -- (-0.439,1.900);
\fill[catP] (0.990,4.287) circle (1.15pt);
\fill[catP] (2.245,3.345) circle (1.15pt);
\fill[catP] (2.969,2.098) circle (1.15pt);
\fill[catP] (3.453,0.885) circle (1.15pt);
\fill[catP] (3.194,-0.364) circle (1.15pt);
\fill[catP] (2.759,-1.429) circle (1.15pt);
\fill[catP] (2.155,-2.414) circle (1.15pt);
\fill[catP] (1.008,-2.643) circle (1.15pt);
\fill[catP] (0.000,-2.886) circle (1.15pt);
\fill[catP] (-1.072,-2.810) circle (1.15pt);
\fill[catP] (-1.750,-1.961) circle (1.15pt);
\fill[catP] (-2.315,-1.199) circle (1.15pt);
\fill[catP] (-2.782,-0.317) circle (1.15pt);
\fill[catP] (-2.484,0.637) circle (1.15pt);
\fill[catP] (-1.954,1.381) circle (1.15pt);
\fill[catP] (-1.218,1.815) circle (1.15pt);
\fill[catP] (-0.439,1.900) circle (1.15pt);
\draw[catR, line width=0.9pt, line join=round] (0.881,3.814) -- (1.947,2.900) -- (2.514,1.777) -- (2.913,0.747) -- (2.825,-0.322) -- (2.049,-1.061) -- (1.541,-1.727) -- (0.853,-2.236) -- (0.000,-2.057) -- (-0.802,-2.102) -- (-1.298,-1.455) -- (-1.725,-0.893) -- (-2.065,-0.235) -- (-2.069,0.530) -- (-1.493,1.055) -- (-0.864,1.287) -- (-0.331,1.434);
\fill[catR] (0.881,3.814) circle (1.15pt);
\fill[catR] (1.947,2.900) circle (1.15pt);
\fill[catR] (2.514,1.777) circle (1.15pt);
\fill[catR] (2.913,0.747) circle (1.15pt);
\fill[catR] (2.825,-0.322) circle (1.15pt);
\fill[catR] (2.049,-1.061) circle (1.15pt);
\fill[catR] (1.541,-1.727) circle (1.15pt);
\fill[catR] (0.853,-2.236) circle (1.15pt);
\fill[catR] (0.000,-2.057) circle (1.15pt);
\fill[catR] (-0.802,-2.102) circle (1.15pt);
\fill[catR] (-1.298,-1.455) circle (1.15pt);
\fill[catR] (-1.725,-0.893) circle (1.15pt);
\fill[catR] (-2.065,-0.235) circle (1.15pt);
\fill[catR] (-2.069,0.530) circle (1.15pt);
\fill[catR] (-1.493,1.055) circle (1.15pt);
\fill[catR] (-0.864,1.287) circle (1.15pt);
\fill[catR] (-0.331,1.434) circle (1.15pt);
\draw[catS, line width=0.9pt, line join=round] (1.112,4.816) -- (2.663,3.967) -- (3.442,2.432) -- (3.716,0.952) -- (3.314,-0.378) -- (2.404,-1.245) -- (1.508,-1.689) -- (0.955,-2.503) -- (0.000,-2.429) -- (-0.646,-1.695) -- (-1.698,-1.902) -- (-1.916,-0.992) -- (-1.568,-0.179) -- (-1.834,0.470) -- (-1.517,1.072) -- (-1.035,1.542) -- (-0.305,1.322);
\fill[catS] (1.112,4.816) circle (1.15pt);
\fill[catS] (2.663,3.967) circle (1.15pt);
\fill[catS] (3.442,2.432) circle (1.15pt);
\fill[catS] (3.716,0.952) circle (1.15pt);
\fill[catS] (3.314,-0.378) circle (1.15pt);
\fill[catS] (2.404,-1.245) circle (1.15pt);
\fill[catS] (1.508,-1.689) circle (1.15pt);
\fill[catS] (0.955,-2.503) circle (1.15pt);
\fill[catS] (0.000,-2.429) circle (1.15pt);
\fill[catS] (-0.646,-1.695) circle (1.15pt);
\fill[catS] (-1.698,-1.902) circle (1.15pt);
\fill[catS] (-1.916,-0.992) circle (1.15pt);
\fill[catS] (-1.568,-0.179) circle (1.15pt);
\fill[catS] (-1.834,0.470) circle (1.15pt);
\fill[catS] (-1.517,1.072) circle (1.15pt);
\fill[catS] (-1.035,1.542) circle (1.15pt);
\fill[catS] (-0.305,1.322) circle (1.15pt);
\draw[catP, line width=0.9pt] (-4.20,-8.35) -- (-3.50,-8.35);
\fill[catP] (-3.850,-8.350) circle (1.15pt);
\node[anchor=west, font=\sffamily\footnotesize] at (-3.40,-8.35) {Perception};
\draw[catR, line width=0.9pt] (-0.30,-8.35) -- (0.40,-8.35);
\fill[catR] (0.050,-8.350) circle (1.15pt);
\node[anchor=west, font=\sffamily\footnotesize] at (0.50,-8.35) {Reasoning};
\draw[catS, line width=0.9pt] (3.38,-8.35) -- (4.08,-8.35);
\fill[catS] (3.730,-8.350) circle (1.15pt);
\node[anchor=west, font=\sffamily\footnotesize] at (4.18,-8.35) {Insight};
\end{tikzpicture}}
\caption{\textbf{Capability-level scores on \bench{} for selected models} (accuracy, \%).
The plot compares ten closed-source and seven open-weight models across Perception, Reasoning,
and Insight. Each spoke represents a model, numbered and ordered clockwise from the top by
overall accuracy. The radial axis ranges from 20\% at the centre to 90\% at the rim. Results for all 32 models are shown
in \cref{fig:polarall}; exact scores are reported in \cref{tab:full} (\cref{app:results}).}
\label{fig:polar}
\end{figure}

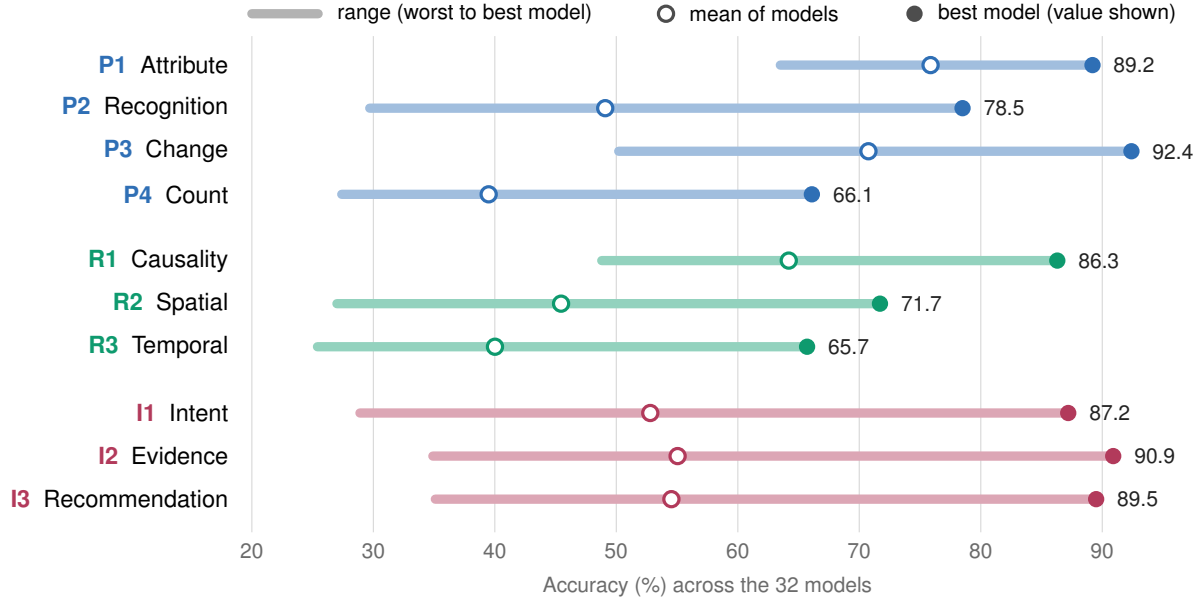
\begin{figure}[t]
\centering
\resizebox{\textwidth}{!}{%
\begin{tikzpicture}
\draw[rule, line width=0.4pt] (0.000,0.35) -- (0.000,-5.64);
\node[font=\sffamily\scriptsize, text=muted, below] at (0.000,-5.64) {20};
\draw[rule, line width=0.4pt] (1.467,0.35) -- (1.467,-5.64);
\node[font=\sffamily\scriptsize, text=muted, below] at (1.467,-5.64) {30};
\draw[rule, line width=0.4pt] (2.933,0.35) -- (2.933,-5.64);
\node[font=\sffamily\scriptsize, text=muted, below] at (2.933,-5.64) {40};
\draw[rule, line width=0.4pt] (4.400,0.35) -- (4.400,-5.64);
\node[font=\sffamily\scriptsize, text=muted, below] at (4.400,-5.64) {50};
\draw[rule, line width=0.4pt] (5.867,0.35) -- (5.867,-5.64);
\node[font=\sffamily\scriptsize, text=muted, below] at (5.867,-5.64) {60};
\draw[rule, line width=0.4pt] (7.333,0.35) -- (7.333,-5.64);
\node[font=\sffamily\scriptsize, text=muted, below] at (7.333,-5.64) {70};
\draw[rule, line width=0.4pt] (8.800,0.35) -- (8.800,-5.64);
\node[font=\sffamily\scriptsize, text=muted, below] at (8.800,-5.64) {80};
\draw[rule, line width=0.4pt] (10.267,0.35) -- (10.267,-5.64);
\node[font=\sffamily\scriptsize, text=muted, below] at (10.267,-5.64) {90};
\node[font=\sffamily\scriptsize, text=muted] at (5.500,-6.29) {Accuracy (\%) across the 32 models};
\node[anchor=east, font=\sffamily\footnotesize] at (-0.15,0.00) {\textbf{\textcolor{catP}{P1}}\; Attribute};
\draw[catP!45, line width=3.2pt, line cap=round] (6.380,0.00) -- (10.149,0.00);
\fill[white] (8.194,0.00) circle (2.6pt); \draw[catP, line width=1.1pt] (8.194,0.00) circle (2.6pt);
\fill[catP] (10.149,0.00) circle (2.8pt);
\node[anchor=west, font=\sffamily\scriptsize, text=ink] at (10.269,0.00) {89.2};
\node[anchor=east, font=\sffamily\footnotesize] at (-0.15,-0.52) {\textbf{\textcolor{catP}{P2}}\; Recognition};
\draw[catP!45, line width=3.2pt, line cap=round] (1.423,-0.52) -- (8.580,-0.52);
\fill[white] (4.266,-0.52) circle (2.6pt); \draw[catP, line width=1.1pt] (4.266,-0.52) circle (2.6pt);
\fill[catP] (8.580,-0.52) circle (2.8pt);
\node[anchor=west, font=\sffamily\scriptsize, text=ink] at (8.700,-0.52) {78.5};
\node[anchor=east, font=\sffamily\footnotesize] at (-0.15,-1.04) {\textbf{\textcolor{catP}{P3}}\; Change};
\draw[catP!45, line width=3.2pt, line cap=round] (4.429,-1.04) -- (10.619,-1.04);
\fill[white] (7.444,-1.04) circle (2.6pt); \draw[catP, line width=1.1pt] (7.444,-1.04) circle (2.6pt);
\fill[catP] (10.619,-1.04) circle (2.8pt);
\node[anchor=west, font=\sffamily\scriptsize, text=ink] at (10.739,-1.04) {92.4};
\node[anchor=east, font=\sffamily\footnotesize] at (-0.15,-1.56) {\textbf{\textcolor{catP}{P4}}\; Count};
\draw[catP!45, line width=3.2pt, line cap=round] (1.085,-1.56) -- (6.761,-1.56);
\fill[white] (2.859,-1.56) circle (2.6pt); \draw[catP, line width=1.1pt] (2.859,-1.56) circle (2.6pt);
\fill[catP] (6.761,-1.56) circle (2.8pt);
\node[anchor=west, font=\sffamily\scriptsize, text=ink] at (6.881,-1.56) {66.1};
\node[anchor=east, font=\sffamily\footnotesize] at (-0.15,-2.36) {\textbf{\textcolor{catR}{R1}}\; Causality};
\draw[catR!45, line width=3.2pt, line cap=round] (4.224,-2.36) -- (9.724,-2.36);
\fill[white] (6.482,-2.36) circle (2.6pt); \draw[catR, line width=1.1pt] (6.482,-2.36) circle (2.6pt);
\fill[catR] (9.724,-2.36) circle (2.8pt);
\node[anchor=west, font=\sffamily\scriptsize, text=ink] at (9.844,-2.36) {86.3};
\node[anchor=east, font=\sffamily\footnotesize] at (-0.15,-2.88) {\textbf{\textcolor{catR}{R2}}\; Spatial};
\draw[catR!45, line width=3.2pt, line cap=round] (1.027,-2.88) -- (7.583,-2.88);
\fill[white] (3.733,-2.88) circle (2.6pt); \draw[catR, line width=1.1pt] (3.733,-2.88) circle (2.6pt);
\fill[catR] (7.583,-2.88) circle (2.8pt);
\node[anchor=west, font=\sffamily\scriptsize, text=ink] at (7.703,-2.88) {71.7};
\node[anchor=east, font=\sffamily\footnotesize] at (-0.15,-3.40) {\textbf{\textcolor{catR}{R3}}\; Temporal};
\draw[catR!45, line width=3.2pt, line cap=round] (0.792,-3.40) -- (6.703,-3.40);
\fill[white] (2.935,-3.40) circle (2.6pt); \draw[catR, line width=1.1pt] (2.935,-3.40) circle (2.6pt);
\fill[catR] (6.703,-3.40) circle (2.8pt);
\node[anchor=west, font=\sffamily\scriptsize, text=ink] at (6.823,-3.40) {65.7};
\node[anchor=east, font=\sffamily\footnotesize] at (-0.15,-4.20) {\textbf{\textcolor{catS}{I1}}\; Intent};
\draw[catS!45, line width=3.2pt, line cap=round] (1.305,-4.20) -- (9.856,-4.20);
\fill[white] (4.810,-4.20) circle (2.6pt); \draw[catS, line width=1.1pt] (4.810,-4.20) circle (2.6pt);
\fill[catS] (9.856,-4.20) circle (2.8pt);
\node[anchor=west, font=\sffamily\scriptsize, text=ink] at (9.976,-4.20) {87.2};
\node[anchor=east, font=\sffamily\footnotesize] at (-0.15,-4.72) {\textbf{\textcolor{catS}{I2}}\; Evidence};
\draw[catS!45, line width=3.2pt, line cap=round] (2.185,-4.72) -- (10.399,-4.72);
\fill[white] (5.140,-4.72) circle (2.6pt); \draw[catS, line width=1.1pt] (5.140,-4.72) circle (2.6pt);
\fill[catS] (10.399,-4.72) circle (2.8pt);
\node[anchor=west, font=\sffamily\scriptsize, text=ink] at (10.519,-4.72) {90.9};
\node[anchor=east, font=\sffamily\footnotesize] at (-0.15,-5.24) {\textbf{\textcolor{catS}{I3}}\; Recommendation};
\draw[catS!45, line width=3.2pt, line cap=round] (2.215,-5.24) -- (10.193,-5.24);
\fill[white] (5.065,-5.24) circle (2.6pt); \draw[catS, line width=1.1pt] (5.065,-5.24) circle (2.6pt);
\fill[catS] (10.193,-5.24) circle (2.8pt);
\node[anchor=west, font=\sffamily\scriptsize, text=ink] at (10.313,-5.24) {89.5};
\draw[black!30, line width=3.2pt, line cap=round] (0.0,0.62) -- (0.8,0.62); \node[anchor=west, font=\sffamily\scriptsize] at (0.9,0.62) {range (worst to best model)};
\fill[white] (5.0,0.62) circle (2.6pt); \draw[black!70, line width=1.1pt] (5.0,0.62) circle (2.6pt); \node[anchor=west, font=\sffamily\scriptsize] at (5.15,0.62) {mean of models};
\fill[black!70] (8.0,0.62) circle (2.8pt); \node[anchor=west, font=\sffamily\scriptsize] at (8.15,0.62) {best model (value shown)};
\end{tikzpicture}}
\caption{\textbf{Task difficulty on \bench{}.} For each task, the bar spans the worst to the best
of the 32 models; the hollow dot is the mean across models. Count (P4) spans 27.4\% to 66.1\% and
Temporal (R3) 25.4\% to 65.7\%, while the three Insight probes have the widest spreads.}
\label{fig:difficulty}
\end{figure}

\subsection{Model performance analysis}
\label{sec:strengths}

\paragraph{Capability-level gaps widen toward Insight.}
The four leading models---Gemini 4 Argon, GPT-6 Astra, GPT-6.1 Sol, and Gemini 3.8 Flash---occupy the top four positions at every capability level. They outperform the other 13 closed-source models by an average of 19.7 percentage points overall, with gaps of 15.6 points on Perception, 18.3 on Reasoning, and 28.9 on Insight. The gap between the best closed-source and open-weight models widens similarly, from 16.3 percentage points on Perception and 22.4 on Reasoning to 37.4 on Insight. Averaged across model groups, the closed-source advantage is 10.3 points on Perception, 9.3 on Reasoning, and 17.1 on Insight. The strongest open-weight model, Muse-Glimmer-30B, ranks ninth overall.

\paragraph{Spatial and temporal reasoning remain challenging, while Insight best separates models.}
Spatial (R2) scores range from 27.0\% to 71.7\% across models (a 44.7 percentage-point
spread; mean 45.5\%), while Temporal (R3) ranges from 25.4\% to 65.7\% (a 40.3-point
spread; mean 40.0\%). Both tasks show substantial variation across models, with Temporal
proving more difficult on average. Insight scores span an even wider range of 55.2 percentage
points (34.0\%--89.2\%), and all three Insight tasks have wider spreads (54.4--58.3 points)
than any Perception or Reasoning task. Insight therefore most strongly differentiates model
performance.

\paragraph{Scale and model development yield uneven gains.}
Model size alone produces limited improvements: across the Qwen3.5 family, a roughly 44$\times$ increase in model size yields a gain of only about 4 percentage points. By contrast, successive evaluated Gemini Flash releases improve overall accuracy by 11.4 and 8.1 points, and Gemini 3.8 Flash outperforms Gemini 3.1 Pro. The evaluated MiMo-V2.6 post-training variants show smaller differences: the MOPD and reinforcement learning (RL) variants score within one percentage point of each other at both model scales.

\paragraph{Capability profiles reveal complementary strengths and weaknesses.}
Insight separates models more strongly than the other levels: its score spread is 55.2 percentage points, compared with 35.1 for Perception and 36.5 for Reasoning. All three Insight tasks also show wider spreads (54.4--58.3 points) than any Perception or Reasoning task. Among the top four models, Insight is the strongest level, scoring 7.6--10.5 points above Perception; among the remaining 28 models, 24 score lower on Insight than on Perception. This pattern is especially pronounced for open-weight models: 14 of 15 score lower on Insight than on Perception, by 8.4 points on average, compared with 1.7 points for closed-source models. However, the pattern is not universal. GPT-4o, Gemini 2.5 Flash, and Qwen3-VL-8B Instruct rank substantially higher on Insight (11th, 12th, and 14th, respectively) than on Perception (18th, 15th, and 29th) or Reasoning (17th, 32nd, and 30th). Their relative strength in subject interpretation and homeowner recommendations, despite weaker performance on visual details and event order, illustrates why \bench{} evaluates the three capability levels separately.
\section{Conclusion}
\label{sec:conclusion}

We introduced \bench{}, a benchmark of 2,254 questions on 1,118 real smart-home camera clips,
designed to assess whether VLMs can understand \emph{what happened and whether it matters}
under practical viewing conditions. Its 10 tasks span three capability levels: Perception
assesses visible content and changes; Reasoning assesses spatial, temporal and causal
relationships; and Insight assesses subject intent, supporting evidence and appropriate
homeowner responses. Across 32 models, the best achieves 81.2\% overall accuracy. Counting
and temporal ordering remain challenging even for the strongest model, while the gap between
the best closed-source and open-weight models is largest on Insight, at 37.4 percentage
points. These findings highlight the difficulty of moving from recognising visual details
and relating events to interpreting subject intent and determining what the available
evidence supports. By scoring each capability level separately, \bench{} reveals differences
that an aggregate score alone would obscure. We hope \bench{} supports the development of VLMs 
that can reliably interpret real smart-home footage, including footage with challenging camera 
viewpoints, partial visibility, challenging lighting and incomplete recordings of events.

\bibliographystyle{plainnat}
\bibliography{refs}

\begin{thebibliography}{15}
\providecommand{\natexlab}[1]{#1}
\providecommand{\url}[1]{\texttt{#1}}
\expandafter\ifx\csname urlstyle\endcsname\relax
  \providecommand{\doi}[1]{doi: #1}\else
  \providecommand{\doi}{doi: \begingroup \urlstyle{rm}\Url}\fi

\bibitem[Fu et~al.(2025)Fu, Dai, Luo, Li, Ren, Zhang, Wang, Zhou, Shen, Zhang, et~al.]{fu2025videomme}
Chaoyou Fu, Yuhan Dai, Yongdong Luo, Lei Li, Shuhuai Ren, Renrui Zhang, Zihan Wang, Chenyu Zhou, Yunhang Shen, Mengdan Zhang, et~al.
\newblock Video-{MME}: The first-ever comprehensive evaluation benchmark of multi-modal {LLMs} in video analysis.
\newblock In \emph{Proceedings of the IEEE/CVF Conference on Computer Vision and Pattern Recognition (CVPR)}, 2025.

\bibitem[Fu et~al.(2026)Fu, Yuan, Dong, Zhang, Shen, Hu, Li, Su, Long, Xie, Xie, Zheng, Yang, Cao, Wu, Liu, Sun, Shan, and He]{fu2026videommev2}
Chaoyou Fu, Haozhi Yuan, Yuhao Dong, Yi-Fan Zhang, Yunhang Shen, Xiaoxing Hu, Xueying Li, Jinsen Su, Chengwu Long, Xiaoyao Xie, Yongkang Xie, Xiawu Zheng, Xue Yang, Haoyu Cao, Yunsheng Wu, Ziwei Liu, Xing Sun, Caifeng Shan, and Ran He.
\newblock {Video-MME-v2}: Towards the next stage in benchmarks for comprehensive video understanding.
\newblock \emph{arXiv preprint arXiv:2604.05015}, 2026.

\bibitem[Li et~al.(2024)Li, Wang, He, Li, Wang, Liu, Wang, Xu, Chen, Luo, Wang, and Qiao]{li2024mvbench}
Kunchang Li, Yali Wang, Yinan He, Yizhuo Li, Yi~Wang, Yi~Liu, Zun Wang, Jilan Xu, Guo Chen, Ping Luo, Limin Wang, and Yu~Qiao.
\newblock {MVBench}: A comprehensive multi-modal video understanding benchmark.
\newblock In \emph{Proceedings of the IEEE/CVF Conference on Computer Vision and Pattern Recognition (CVPR)}, 2024.

\bibitem[Liu et~al.(2024)Liu, Li, Liu, Wang, Ren, Li, Chen, Sun, and Hou]{liu2024tempcompass}
Yuanxin Liu, Shicheng Li, Yi~Liu, Yuxiang Wang, Shuhuai Ren, Lei Li, Sishuo Chen, Xu~Sun, and Lu~Hou.
\newblock {TempCompass}: Do video {LLMs} really understand videos?
\newblock In \emph{Findings of the Association for Computational Linguistics (ACL)}, 2024.

\bibitem[Mangalam et~al.(2023)Mangalam, Akshulakov, and Malik]{mangalam2023egoschema}
Karttikeya Mangalam, Raiymbek Akshulakov, and Jitendra Malik.
\newblock {EgoSchema}: A diagnostic benchmark for very long-form video language understanding.
\newblock In \emph{Advances in Neural Information Processing Systems (NeurIPS), Datasets and Benchmarks Track}, 2023.

\bibitem[P{\u{a}}tr{\u{a}}ucean et~al.(2023)P{\u{a}}tr{\u{a}}ucean, Smaira, Gupta, Recasens, Markeeva, Banarse, Koppula, Heyward, Malinowski, Yang, et~al.]{patraucean2023perception}
Viorica P{\u{a}}tr{\u{a}}ucean, Lucas Smaira, Ankush Gupta, Adri{\`a} Recasens, Larisa Markeeva, Dylan Banarse, Skanda Koppula, Joseph Heyward, Mateusz Malinowski, Yi~Yang, et~al.
\newblock Perception test: A diagnostic benchmark for multimodal video models.
\newblock In \emph{Advances in Neural Information Processing Systems (NeurIPS), Datasets and Benchmarks Track}, 2023.

\bibitem[Sigurdsson et~al.(2016)Sigurdsson, Varol, Wang, Farhadi, Laptev, and Gupta]{sigurdsson2016charades}
Gunnar~A. Sigurdsson, G{\"u}l Varol, Xiaolong Wang, Ali Farhadi, Ivan Laptev, and Abhinav Gupta.
\newblock Hollywood in homes: Crowdsourcing data collection for activity understanding.
\newblock In \emph{European Conference on Computer Vision (ECCV)}, 2016.

\bibitem[Sultani et~al.(2018)Sultani, Chen, and Shah]{sultani2018ucfcrime}
Waqas Sultani, Chen Chen, and Mubarak Shah.
\newblock Real-world anomaly detection in surveillance videos.
\newblock In \emph{Proceedings of the IEEE Conference on Computer Vision and Pattern Recognition (CVPR)}, 2018.

\bibitem[Wu et~al.(2024)Wu, Li, Chen, and Li]{wu2024longvideobench}
Haoning Wu, Dongxu Li, Bei Chen, and Junnan Li.
\newblock {LongVideoBench}: A benchmark for long-context interleaved video-language understanding.
\newblock In \emph{Advances in Neural Information Processing Systems (NeurIPS), Datasets and Benchmarks Track}, 2024.

\bibitem[Xiao et~al.(2021)Xiao, Shang, Yao, and Chua]{xiao2021nextqa}
Junbin Xiao, Xindi Shang, Angela Yao, and Tat-Seng Chua.
\newblock {NExT-QA}: Next phase of question-answering to explaining temporal actions.
\newblock In \emph{Proceedings of the IEEE/CVF Conference on Computer Vision and Pattern Recognition (CVPR)}, 2021.

\bibitem[Yi et~al.(2020)Yi, Gan, Li, Kohli, Wu, Torralba, and Tenenbaum]{yi2020clevrer}
Kexin Yi, Chuang Gan, Yunzhu Li, Pushmeet Kohli, Jiajun Wu, Antonio Torralba, and Joshua~B. Tenenbaum.
\newblock {CLEVRER}: Collision events for video representation and reasoning.
\newblock In \emph{International Conference on Learning Representations (ICLR)}, 2020.

\bibitem[Yu et~al.(2019)Yu, Xu, Yu, Yu, Zhao, Zhuang, and Tao]{yu2019activitynetqa}
Zhou Yu, Dejing Xu, Jun Yu, Ting Yu, Zhou Zhao, Yueting Zhuang, and Dacheng Tao.
\newblock {ActivityNet-QA}: A dataset for understanding complex web videos via question answering.
\newblock In \emph{Proceedings of the AAAI Conference on Artificial Intelligence}, 2019.

\bibitem[Zadeh et~al.(2019)Zadeh, Chan, Liang, Tong, and Morency]{zadeh2019socialiq}
Amir Zadeh, Michael Chan, Paul~Pu Liang, Edmund Tong, and Louis-Philippe Morency.
\newblock Social-{IQ}: A question answering benchmark for artificial social intelligence.
\newblock In \emph{Proceedings of the IEEE/CVF Conference on Computer Vision and Pattern Recognition (CVPR)}, 2019.

\bibitem[Zhao et~al.(2025)Zhao, Zhang, Guo, Li, Chen, Zhao, and Huang]{zhao2025smarthomebench}
Xinyi Zhao, Congjing Zhang, Pei Guo, Wei Li, Lin Chen, Chaoyue Zhao, and Shuai Huang.
\newblock {SmartHome-Bench}: A comprehensive benchmark for video anomaly detection in smart homes using multi-modal large language models.
\newblock In \emph{Proceedings of the IEEE/CVF Conference on Computer Vision and Pattern Recognition (CVPR) Workshops}, pages 4014--4024, June 2025.

\bibitem[Zhou et~al.(2024)Zhou, Shu, Zhao, Wu, Xiao, Yang, Xiong, Zhang, Huang, and Liu]{zhou2024mlvu}
Junjie Zhou, Yan Shu, Bo~Zhao, Boya Wu, Shitao Xiao, Xi~Yang, Yongping Xiong, Bo~Zhang, Tiejun Huang, and Zheng Liu.
\newblock {MLVU}: A comprehensive benchmark for multi-task long video understanding.
\newblock \emph{arXiv preprint arXiv:2406.04264}, 2024.

\end{thebibliography}

\appendix
\section{Full Results and Model Configurations}
\label{app:results}
\begin{sidewaystable}
\centering
\renewcommand{\arraystretch}{1.12}
\setlength{\tabcolsep}{4pt}
\caption{\textbf{Full results and settings on \bench{}} (accuracy, \%), sorted by overall accuracy. Input:
\emph{frames}, frames sent through the provider's API; \emph{video}, the video file served locally
with SGLang or vLLM; both follow the sampling protocol in \cref{sec:protocol}. Unless specified in \cref{sec:protocol} or this table, settings use
provider or serving-framework defaults. Perc., Reas. and Ins. denote Perception, Reasoning and Insight, respectively. Level scores are accuracy over the questions at each level, and Overall is
accuracy over all 2,254 questions; these are the values plotted in \cref{fig:polarall}. Best per
column in \textbf{bold}, second \underline{underlined}.}
\label{tab:full}
\resizebox{\textheight}{!}{%
\begin{tabular}{@{}l l l *{4}{c} | *{3}{c} | *{3}{c} | c c c | c@{}}
\toprule
& & & \multicolumn{4}{c|}{\textbf{Perception}} & \multicolumn{3}{c|}{\textbf{Reasoning}} & \multicolumn{3}{c|}{\textbf{Insight}} & \multicolumn{3}{c|}{\textbf{Level}} & \\
\textbf{Model} & \textbf{Input} & \textbf{Settings} & P1 & P2 & P3 & P4 & R1 & R2 & R3 & I1 & I2 & I3 & \textbf{Perc.} & \textbf{Reas.} & \textbf{Ins.} & \textbf{Overall} \\
\midrule
Gemini 4 Argon & frames & \texttt{thinking\_level low} & \textbf{89.2} & \textbf{78.5} & \textbf{92.4} & \textbf{66.1} & \textbf{86.3} & \textbf{71.7} & \textbf{65.7} & \textbf{87.2} & \textbf{90.9} & \textbf{89.5} & \textbf{81.6} & \textbf{74.8} & \textbf{89.2} & \textbf{81.2} \\
GPT-6 Astra & frames & \texttt{reasoning\_effort low} & 83.9 & \underline{72.8} & 87.2 & \underline{61.6} & \underline{79.0} & \underline{70.5} & \underline{56.8} & \underline{84.5} & \underline{89.8} & \underline{86.5} & \underline{76.4} & \underline{68.9} & \underline{86.9} & \underline{76.6} \\
GPT-6.1 Sol & frames & \texttt{reasoning\_effort low} & 82.7 & 68.7 & \underline{87.6} & 55.9 & 78.2 & 67.9 & 54.7 & 78.1 & 86.6 & 80.1 & 73.8 & 67.1 & 81.6 & 73.6 \\
Gemini 3.8 Flash & frames & \texttt{thinking\_level low} & 82.3 & 65.5 & 84.7 & 50.6 & 75.8 & 61.6 & 51.3 & 80.2 & 79.6 & 77.2 & 70.9 & 63.1 & 79.0 & 70.4 \\
Muse Wonder 1.0 & frames & \texttt{low resolution, low reasoning} & \underline{84.3} & 61.4 & 81.9 & 51.4 & 73.0 & 63.3 & 49.6 & 68.4 & 78.5 & 74.3 & 69.9 & 62.1 & 73.7 & 68.3 \\
Gemini 3.1 Pro & frames & \texttt{thinking\_level low} & 79.9 & 58.5 & 75.1 & 46.1 & 73.4 & 59.5 & 45.8 & 64.7 & 68.8 & 66.7 & 65.0 & 59.8 & 66.7 & 63.8 \\
Gemini 3.5 Flash & frames & \texttt{thinking\_level minimal} & 83.1 & 61.8 & 73.5 & 46.5 & 66.5 & 53.6 & 44.5 & 61.0 & 68.3 & 64.3 & 66.3 & 55.1 & 64.5 & 62.3 \\
Gemini 2.5 Pro & frames & \texttt{thinking\_budget 128} & 80.3 & 59.4 & 73.5 & 40.4 & 66.9 & 46.0 & 43.2 & 52.9 & 59.7 & 61.4 & 63.5 & 52.3 & 57.9 & 58.6 \\
Muse-Glimmer-30B & frames & \texttt{reasoning\_strength low} & 81.9 & 55.3 & 77.1 & 46.5 & 63.3 & 50.2 & 43.2 & 48.1 & 52.2 & 55.0 & 65.3 & 52.4 & 51.7 & 57.9 \\
GPT-6 Luna & frames & \texttt{reasoning\_effort none} & 78.7 & 46.7 & 73.9 & 38.0 & 66.5 & 48.9 & 43.6 & 55.1 & 67.2 & 57.9 & 59.4 & 53.2 & 60.1 & 57.6 \\
Grok 4.7 & frames & \texttt{reasoning\_effort minimal, detail low} & 77.1 & 45.9 & 74.7 & 40.4 & 64.1 & 50.6 & 45.3 & 50.8 & 71.0 & 50.3 & 59.6 & 53.5 & 57.5 & 57.2 \\
Gemini 3.5 Flash-Lite & frames & \texttt{thinking\_level minimal} & 77.1 & 55.3 & 65.9 & 42.9 & 67.7 & 43.0 & 34.8 & 51.3 & 54.3 & 56.7 & 60.4 & 48.8 & 54.0 & 55.2 \\
Gemma 4 31B-it & frames & \texttt{no thinking, temperature 0} & 78.7 & 45.9 & 79.9 & 43.3 & 63.7 & 47.7 & 42.4 & 46.0 & 44.1 & 46.1 & 62.1 & 51.5 & 45.4 & 54.7 \\
GPT-4o & frames & \texttt{temperature 0} & 74.3 & 52.8 & 66.7 & 33.1 & 65.7 & 37.1 & 38.1 & 51.3 & 60.8 & 55.0 & 56.8 & 47.3 & 55.7 & 53.5 \\
Qwen3.5-397B-A17B & video & \texttt{SGLang} & 76.3 & 44.7 & 71.9 & 32.7 & 62.1 & 38.4 & 40.3 & 45.5 & 52.7 & 52.6 & 56.5 & 47.2 & 50.2 & 52.0 \\
GPT-5 mini & frames & \texttt{reasoning\_effort minimal} & 79.9 & 54.5 & 67.9 & 33.9 & 60.1 & 45.6 & 41.1 & 45.5 & 38.2 & 42.7 & 59.2 & 49.1 & 42.1 & 51.8 \\
GPT-4.1 & frames & \texttt{temperature 0} & 75.5 & 53.7 & 63.9 & 33.9 & 61.7 & 42.6 & 35.2 & 48.7 & 48.9 & 50.9 & 56.9 & 46.7 & 49.4 & 51.8 \\
Qwen3.8-Flash-Next & video & \texttt{SGLang} & 73.5 & 43.1 & 71.5 & 35.1 & 65.7 & 45.1 & 38.1 & 50.8 & 45.7 & 42.7 & 55.9 & 49.9 & 46.5 & 51.7 \\
Qwen3.8-27B & video & \texttt{SGLang} & 69.9 & 37.0 & 72.3 & 32.2 & 66.1 & 38.4 & 42.8 & 50.3 & 48.9 & 49.1 & 53.0 & 49.4 & 49.4 & 51.0 \\
Gemini 2.5 Flash & frames & \texttt{thinking\_budget 0} & 81.5 & 52.0 & 70.3 & 29.0 & 59.3 & 27.0 & 27.5 & 56.7 & 48.9 & 57.3 & 58.3 & 38.3 & 54.2 & 50.9 \\
Qwen3.5-122B-A10B & video & \texttt{SGLang} & 75.5 & 41.9 & 73.9 & 36.3 & 59.3 & 39.7 & 39.0 & 42.2 & 45.7 & 46.2 & 57.0 & 46.2 & 44.7 & 50.6 \\
Qwen3.6-35B-A3B & video & \texttt{SGLang} & 75.1 & 38.6 & 69.9 & 36.7 & 61.3 & 40.1 & 37.3 & 44.4 & 41.4 & 48.5 & 55.2 & 46.5 & 44.7 & 49.9 \\
Qwen3.5-27B & video & \texttt{SGLang} & 70.3 & 40.2 & 68.7 & 34.3 & 62.9 & 38.8 & 34.3 & 44.4 & 48.9 & 44.4 & 53.5 & 45.6 & 46.0 & 49.2 \\
Gemini 2.5 Flash-Lite & frames & \texttt{thinking\_budget 0} & 76.7 & 51.2 & 57.4 & 36.7 & 60.9 & 32.9 & 30.1 & 43.9 & 36.6 & 50.9 & 55.6 & 41.6 & 43.6 & 48.2 \\
Qwen3.5-9B & video & \texttt{SGLang} & 69.1 & 35.8 & 66.7 & 35.1 & 58.9 & 38.0 & 38.6 & 39.6 & 46.2 & 46.2 & 51.8 & 45.4 & 43.9 & 47.8 \\
MiMo-V2.6-Pro-MOPD & video & \texttt{vLLM} & 66.7 & 32.5 & 67.5 & 35.1 & 55.6 & 35.4 & 33.5 & 48.1 & 47.8 & 41.5 & 50.6 & 41.7 & 46.0 & 46.6 \\
Qwen3-VL-8B Instruct & video & \texttt{SGLang} & 66.3 & 37.0 & 61.5 & 27.4 & 55.2 & 32.9 & 30.5 & 48.7 & 45.7 & 62.0 & 48.2 & 39.8 & 51.8 & 46.4 \\
MiMo-V2.6-Pro-RL & video & \texttt{vLLM} & 67.1 & 32.5 & 64.7 & 36.3 & 55.6 & 36.3 & 31.8 & 46.5 & 45.2 & 40.4 & 50.3 & 41.5 & 44.1 & 46.0 \\
GPT-4.1 mini & frames & \texttt{temperature 0} & 73.1 & 49.2 & 50.2 & 27.8 & 58.5 & 39.7 & 34.3 & 40.6 & 36.6 & 41.5 & 50.2 & 44.4 & 39.5 & 45.8 \\
MiMo-V2.6-Flash-MOPD & video & \texttt{vLLM} & 66.3 & 34.6 & 55.4 & 32.7 & 55.2 & 39.7 & 26.3 & 41.7 & 39.2 & 35.7 & 47.3 & 40.6 & 39.0 & 43.2 \\
MiMo-V2.6-Flash-RL & video & \texttt{vLLM} & 67.9 & 34.1 & 56.6 & 33.1 & 56.9 & 35.9 & 25.4 & 43.3 & 34.9 & 36.3 & 48.0 & 39.7 & 38.2 & 43.0 \\
MiMo-V2.6-Distill-Qwen-9B & video & \texttt{SGLang} & 63.5 & 29.7 & 59.8 & 32.7 & 48.8 & 36.3 & 35.2 & 28.9 & 38.2 & 35.1 & 46.5 & 40.2 & 34.0 & 41.5 \\
\bottomrule
\end{tabular}}
\end{sidewaystable}

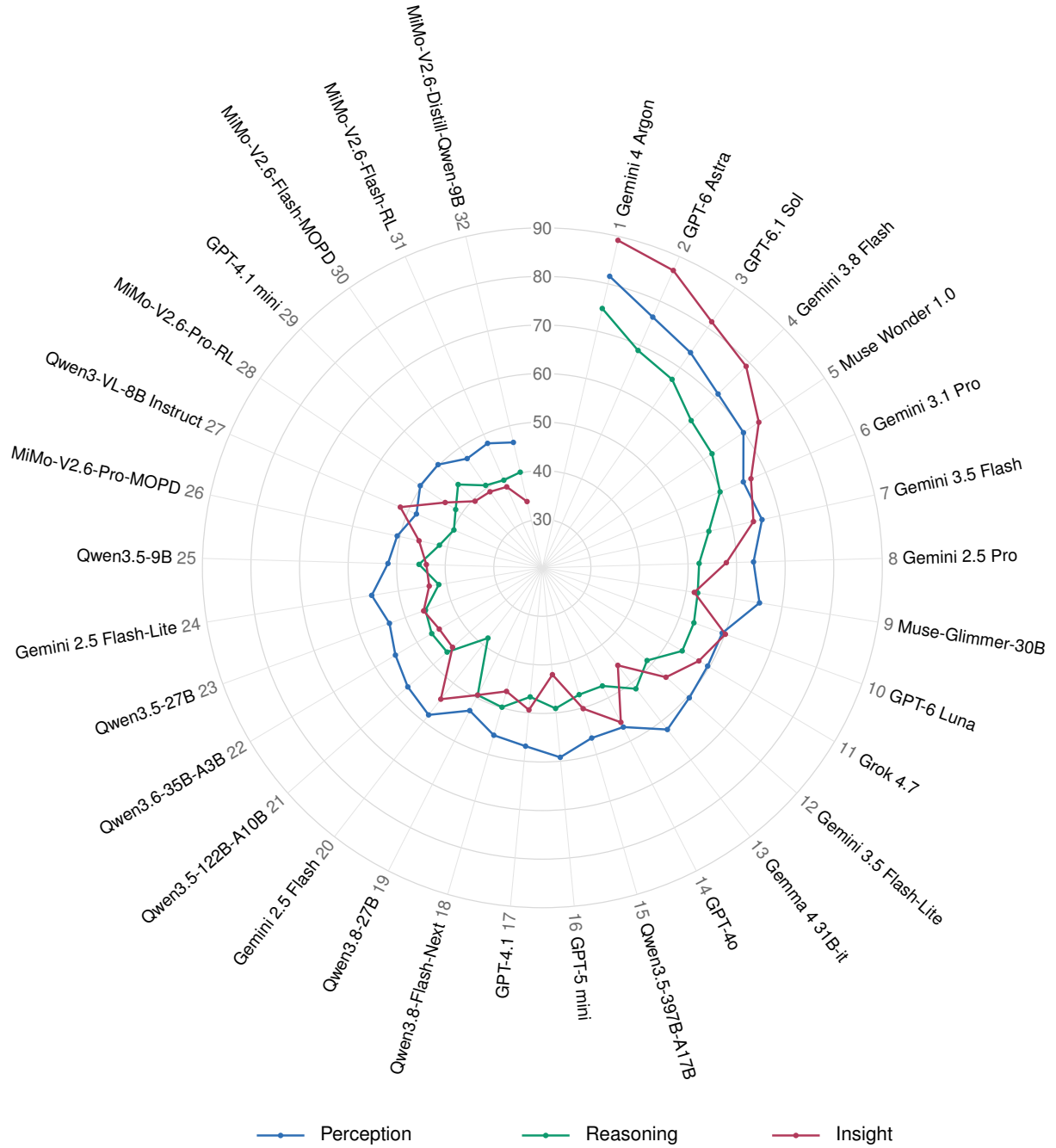
\begin{figure}[p]
\centering
\resizebox{\textwidth}{!}{%
\begin{tikzpicture}
\draw[rule, line width=0.45pt] (0,0) circle (0.714);
\node[font=\sffamily\scriptsize, text=muted, fill=white, inner sep=0.8pt] at (0,0.714) {30};
\draw[rule, line width=0.45pt] (0,0) circle (1.429);
\node[font=\sffamily\scriptsize, text=muted, fill=white, inner sep=0.8pt] at (0,1.429) {40};
\draw[rule, line width=0.45pt] (0,0) circle (2.143);
\node[font=\sffamily\scriptsize, text=muted, fill=white, inner sep=0.8pt] at (0,2.143) {50};
\draw[rule, line width=0.45pt] (0,0) circle (2.857);
\node[font=\sffamily\scriptsize, text=muted, fill=white, inner sep=0.8pt] at (0,2.857) {60};
\draw[rule, line width=0.45pt] (0,0) circle (3.571);
\node[font=\sffamily\scriptsize, text=muted, fill=white, inner sep=0.8pt] at (0,3.571) {70};
\draw[rule, line width=0.45pt] (0,0) circle (4.286);
\node[font=\sffamily\scriptsize, text=muted, fill=white, inner sep=0.8pt] at (0,4.286) {80};
\draw[rule, line width=0.45pt] (0,0) circle (5.000);
\node[font=\sffamily\scriptsize, text=muted, fill=white, inner sep=0.8pt] at (0,5.000) {90};
\draw[rule!70, line width=0.3pt] (0,0) -- (1.125,4.872);
\node[anchor=west, rotate=77.00, font=\sffamily\scriptsize, inner sep=0pt] at (1.144,4.955) {\textcolor{muted}{1}\;Gemini 4 Argon};
\draw[rule!70, line width=0.3pt] (0,0) -- (2.016,4.576);
\node[anchor=west, rotate=66.23, font=\sffamily\scriptsize, inner sep=0pt] at (2.050,4.654) {\textcolor{muted}{2}\;GPT-6 Astra};
\draw[rule!70, line width=0.3pt] (0,0) -- (2.836,4.118);
\node[anchor=west, rotate=55.45, font=\sffamily\scriptsize, inner sep=0pt] at (2.884,4.189) {\textcolor{muted}{3}\;GPT-6.1 Sol};
\draw[rule!70, line width=0.3pt] (0,0) -- (3.555,3.516);
\node[anchor=west, rotate=44.68, font=\sffamily\scriptsize, inner sep=0pt] at (3.616,3.576) {\textcolor{muted}{4}\;Gemini 3.8 Flash};
\draw[rule!70, line width=0.3pt] (0,0) -- (4.150,2.789);
\node[anchor=west, rotate=33.90, font=\sffamily\scriptsize, inner sep=0pt] at (4.221,2.837) {\textcolor{muted}{5}\;Muse Wonder 1.0};
\draw[rule!70, line width=0.3pt] (0,0) -- (4.598,1.964);
\node[anchor=west, rotate=23.13, font=\sffamily\scriptsize, inner sep=0pt] at (4.677,1.998) {\textcolor{muted}{6}\;Gemini 3.1 Pro};
\draw[rule!70, line width=0.3pt] (0,0) -- (4.884,1.070);
\node[anchor=west, rotate=12.35, font=\sffamily\scriptsize, inner sep=0pt] at (4.968,1.088) {\textcolor{muted}{7}\;Gemini 3.5 Flash};
\draw[rule!70, line width=0.3pt] (0,0) -- (4.998,0.138);
\node[anchor=west, rotate=1.58, font=\sffamily\scriptsize, inner sep=0pt] at (5.084,0.140) {\textcolor{muted}{8}\;Gemini 2.5 Pro};
\draw[rule!70, line width=0.3pt] (0,0) -- (4.936,-0.799);
\node[anchor=west, rotate=-9.19, font=\sffamily\scriptsize, inner sep=0pt] at (5.020,-0.813) {\textcolor{muted}{9}\;Muse-Glimmer-30B};
\draw[rule!70, line width=0.3pt] (0,0) -- (4.699,-1.707);
\node[anchor=west, rotate=-19.97, font=\sffamily\scriptsize, inner sep=0pt] at (4.780,-1.737) {\textcolor{muted}{10}\;GPT-6 Luna};
\draw[rule!70, line width=0.3pt] (0,0) -- (4.297,-2.556);
\node[anchor=west, rotate=-30.74, font=\sffamily\scriptsize, inner sep=0pt] at (4.371,-2.600) {\textcolor{muted}{11}\;Grok 4.7};
\draw[rule!70, line width=0.3pt] (0,0) -- (3.744,-3.314);
\node[anchor=west, rotate=-41.52, font=\sffamily\scriptsize, inner sep=0pt] at (3.808,-3.371) {\textcolor{muted}{12}\;Gemini 3.5 Flash-Lite};
\draw[rule!70, line width=0.3pt] (0,0) -- (3.058,-3.956);
\node[anchor=west, rotate=-52.29, font=\sffamily\scriptsize, inner sep=0pt] at (3.111,-4.023) {\textcolor{muted}{13}\;Gemma 4 31B-it};
\draw[rule!70, line width=0.3pt] (0,0) -- (2.265,-4.458);
\node[anchor=west, rotate=-63.06, font=\sffamily\scriptsize, inner sep=0pt] at (2.304,-4.534) {\textcolor{muted}{14}\;GPT-4o};
\draw[rule!70, line width=0.3pt] (0,0) -- (1.392,-4.802);
\node[anchor=west, rotate=-73.84, font=\sffamily\scriptsize, inner sep=0pt] at (1.416,-4.885) {\textcolor{muted}{15}\;Qwen3.5-397B-A17B};
\draw[rule!70, line width=0.3pt] (0,0) -- (0.469,-4.978);
\node[anchor=west, rotate=-84.61, font=\sffamily\scriptsize, inner sep=0pt] at (0.477,-5.063) {\textcolor{muted}{16}\;GPT-5 mini};
\draw[rule!70, line width=0.3pt] (0,0) -- (-0.469,-4.978);
\node[anchor=east, rotate=84.61, font=\sffamily\scriptsize, inner sep=0pt] at (-0.477,-5.063) {GPT-4.1\;\textcolor{muted}{17}};
\draw[rule!70, line width=0.3pt] (0,0) -- (-1.392,-4.802);
\node[anchor=east, rotate=73.84, font=\sffamily\scriptsize, inner sep=0pt] at (-1.416,-4.885) {Qwen3.8-Flash-Next\;\textcolor{muted}{18}};
\draw[rule!70, line width=0.3pt] (0,0) -- (-2.265,-4.458);
\node[anchor=east, rotate=63.06, font=\sffamily\scriptsize, inner sep=0pt] at (-2.304,-4.534) {Qwen3.8-27B\;\textcolor{muted}{19}};
\draw[rule!70, line width=0.3pt] (0,0) -- (-3.058,-3.956);
\node[anchor=east, rotate=52.29, font=\sffamily\scriptsize, inner sep=0pt] at (-3.111,-4.023) {Gemini 2.5 Flash\;\textcolor{muted}{20}};
\draw[rule!70, line width=0.3pt] (0,0) -- (-3.744,-3.314);
\node[anchor=east, rotate=41.52, font=\sffamily\scriptsize, inner sep=0pt] at (-3.808,-3.371) {Qwen3.5-122B-A10B\;\textcolor{muted}{21}};
\draw[rule!70, line width=0.3pt] (0,0) -- (-4.297,-2.556);
\node[anchor=east, rotate=30.74, font=\sffamily\scriptsize, inner sep=0pt] at (-4.371,-2.600) {Qwen3.6-35B-A3B\;\textcolor{muted}{22}};
\draw[rule!70, line width=0.3pt] (0,0) -- (-4.699,-1.707);
\node[anchor=east, rotate=19.97, font=\sffamily\scriptsize, inner sep=0pt] at (-4.780,-1.737) {Qwen3.5-27B\;\textcolor{muted}{23}};
\draw[rule!70, line width=0.3pt] (0,0) -- (-4.936,-0.799);
\node[anchor=east, rotate=9.19, font=\sffamily\scriptsize, inner sep=0pt] at (-5.020,-0.813) {Gemini 2.5 Flash-Lite\;\textcolor{muted}{24}};
\draw[rule!70, line width=0.3pt] (0,0) -- (-4.998,0.138);
\node[anchor=east, rotate=-1.58, font=\sffamily\scriptsize, inner sep=0pt] at (-5.084,0.140) {Qwen3.5-9B\;\textcolor{muted}{25}};
\draw[rule!70, line width=0.3pt] (0,0) -- (-4.884,1.070);
\node[anchor=east, rotate=-12.35, font=\sffamily\scriptsize, inner sep=0pt] at (-4.968,1.088) {MiMo-V2.6-Pro-MOPD\;\textcolor{muted}{26}};
\draw[rule!70, line width=0.3pt] (0,0) -- (-4.598,1.964);
\node[anchor=east, rotate=-23.13, font=\sffamily\scriptsize, inner sep=0pt] at (-4.677,1.998) {Qwen3-VL-8B Instruct\;\textcolor{muted}{27}};
\draw[rule!70, line width=0.3pt] (0,0) -- (-4.150,2.789);
\node[anchor=east, rotate=-33.90, font=\sffamily\scriptsize, inner sep=0pt] at (-4.221,2.837) {MiMo-V2.6-Pro-RL\;\textcolor{muted}{28}};
\draw[rule!70, line width=0.3pt] (0,0) -- (-3.555,3.516);
\node[anchor=east, rotate=-44.68, font=\sffamily\scriptsize, inner sep=0pt] at (-3.616,3.576) {GPT-4.1 mini\;\textcolor{muted}{29}};
\draw[rule!70, line width=0.3pt] (0,0) -- (-2.836,4.118);
\node[anchor=east, rotate=-55.45, font=\sffamily\scriptsize, inner sep=0pt] at (-2.884,4.189) {MiMo-V2.6-Flash-MOPD\;\textcolor{muted}{30}};
\draw[rule!70, line width=0.3pt] (0,0) -- (-2.016,4.576);
\node[anchor=east, rotate=-66.23, font=\sffamily\scriptsize, inner sep=0pt] at (-2.050,4.654) {MiMo-V2.6-Flash-RL\;\textcolor{muted}{31}};
\draw[rule!70, line width=0.3pt] (0,0) -- (-1.125,4.872);
\node[anchor=east, rotate=-77.00, font=\sffamily\scriptsize, inner sep=0pt] at (-1.144,4.955) {MiMo-V2.6-Distill-Qwen-9B\;\textcolor{muted}{32}};
\draw[catP, line width=0.9pt, line join=round] (0.990,4.287) -- (1.624,3.687) -- (2.179,3.165) -- (2.585,2.556) -- (2.958,1.988) -- (2.956,1.263) -- (3.231,0.708) -- (3.106,0.086) -- (3.194,-0.517) -- (2.645,-0.961) -- (2.431,-1.446) -- (2.161,-1.913) -- (1.839,-2.379) -- (1.191,-2.343) -- (0.726,-2.504) -- (0.263,-2.788) -- (-0.247,-2.624) -- (-0.714,-2.463) -- (-1.068,-2.101) -- (-1.673,-2.164) -- (-1.979,-1.752) -- (-2.161,-1.285) -- (-2.249,-0.817) -- (-2.510,-0.406) -- (-2.271,0.063) -- (-2.135,0.468) -- (-1.852,0.791) -- (-1.796,1.207) -- (-1.534,1.517) -- (-1.106,1.606) -- (-0.806,1.830) -- (-0.426,1.844);
\fill[catP] (0.990,4.287) circle (1.15pt);
\fill[catP] (1.624,3.687) circle (1.15pt);
\fill[catP] (2.179,3.165) circle (1.15pt);
\fill[catP] (2.585,2.556) circle (1.15pt);
\fill[catP] (2.958,1.988) circle (1.15pt);
\fill[catP] (2.956,1.263) circle (1.15pt);
\fill[catP] (3.231,0.708) circle (1.15pt);
\fill[catP] (3.106,0.086) circle (1.15pt);
\fill[catP] (3.194,-0.517) circle (1.15pt);
\fill[catP] (2.645,-0.961) circle (1.15pt);
\fill[catP] (2.431,-1.446) circle (1.15pt);
\fill[catP] (2.161,-1.913) circle (1.15pt);
\fill[catP] (1.839,-2.379) circle (1.15pt);
\fill[catP] (1.191,-2.343) circle (1.15pt);
\fill[catP] (0.726,-2.504) circle (1.15pt);
\fill[catP] (0.263,-2.788) circle (1.15pt);
\fill[catP] (-0.247,-2.624) circle (1.15pt);
\fill[catP] (-0.714,-2.463) circle (1.15pt);
\fill[catP] (-1.068,-2.101) circle (1.15pt);
\fill[catP] (-1.673,-2.164) circle (1.15pt);
\fill[catP] (-1.979,-1.752) circle (1.15pt);
\fill[catP] (-2.161,-1.285) circle (1.15pt);
\fill[catP] (-2.249,-0.817) circle (1.15pt);
\fill[catP] (-2.510,-0.406) circle (1.15pt);
\fill[catP] (-2.271,0.063) circle (1.15pt);
\fill[catP] (-2.135,0.468) circle (1.15pt);
\fill[catP] (-1.852,0.791) circle (1.15pt);
\fill[catP] (-1.796,1.207) circle (1.15pt);
\fill[catP] (-1.534,1.517) circle (1.15pt);
\fill[catP] (-1.106,1.606) circle (1.15pt);
\fill[catP] (-0.806,1.830) circle (1.15pt);
\fill[catP] (-0.426,1.844) circle (1.15pt);
\draw[catR, line width=0.9pt, line join=round] (0.881,3.814) -- (1.408,3.196) -- (1.908,2.771) -- (2.189,2.165) -- (2.496,1.677) -- (2.614,1.117) -- (2.449,0.536) -- (2.306,0.064) -- (2.285,-0.370) -- (2.229,-0.810) -- (2.057,-1.223) -- (1.540,-1.364) -- (1.376,-1.780) -- (0.883,-1.738) -- (0.541,-1.866) -- (0.195,-2.069) -- (-0.179,-1.899) -- (-0.594,-2.051) -- (-0.951,-1.872) -- (-0.800,-1.034) -- (-1.401,-1.240) -- (-1.627,-0.968) -- (-1.719,-0.624) -- (-1.523,-0.247) -- (-1.814,0.050) -- (-1.514,0.332) -- (-1.301,0.556) -- (-1.275,0.857) -- (-1.239,1.225) -- (-0.834,1.212) -- (-0.567,1.288) -- (-0.325,1.406);
\fill[catR] (0.881,3.814) circle (1.15pt);
\fill[catR] (1.408,3.196) circle (1.15pt);
\fill[catR] (1.908,2.771) circle (1.15pt);
\fill[catR] (2.189,2.165) circle (1.15pt);
\fill[catR] (2.496,1.677) circle (1.15pt);
\fill[catR] (2.614,1.117) circle (1.15pt);
\fill[catR] (2.449,0.536) circle (1.15pt);
\fill[catR] (2.306,0.064) circle (1.15pt);
\fill[catR] (2.285,-0.370) circle (1.15pt);
\fill[catR] (2.229,-0.810) circle (1.15pt);
\fill[catR] (2.057,-1.223) circle (1.15pt);
\fill[catR] (1.540,-1.364) circle (1.15pt);
\fill[catR] (1.376,-1.780) circle (1.15pt);
\fill[catR] (0.883,-1.738) circle (1.15pt);
\fill[catR] (0.541,-1.866) circle (1.15pt);
\fill[catR] (0.195,-2.069) circle (1.15pt);
\fill[catR] (-0.179,-1.899) circle (1.15pt);
\fill[catR] (-0.594,-2.051) circle (1.15pt);
\fill[catR] (-0.951,-1.872) circle (1.15pt);
\fill[catR] (-0.800,-1.034) circle (1.15pt);
\fill[catR] (-1.401,-1.240) circle (1.15pt);
\fill[catR] (-1.627,-0.968) circle (1.15pt);
\fill[catR] (-1.719,-0.624) circle (1.15pt);
\fill[catR] (-1.523,-0.247) circle (1.15pt);
\fill[catR] (-1.814,0.050) circle (1.15pt);
\fill[catR] (-1.514,0.332) circle (1.15pt);
\fill[catR] (-1.301,0.556) circle (1.15pt);
\fill[catR] (-1.275,0.857) circle (1.15pt);
\fill[catR] (-1.239,1.225) circle (1.15pt);
\fill[catR] (-0.834,1.212) circle (1.15pt);
\fill[catR] (-0.567,1.288) circle (1.15pt);
\fill[catR] (-0.325,1.406) circle (1.15pt);
\draw[catS, line width=0.9pt, line join=round] (1.112,4.816) -- (1.926,4.373) -- (2.491,3.618) -- (2.997,2.963) -- (3.184,2.140) -- (3.068,1.310) -- (3.105,0.680) -- (2.706,0.075) -- (2.235,-0.362) -- (2.692,-0.978) -- (2.302,-1.370) -- (1.818,-1.610) -- (1.110,-1.435) -- (1.155,-2.273) -- (0.600,-2.072) -- (0.148,-1.572) -- (-0.197,-2.091) -- (-0.527,-1.818) -- (-0.951,-1.872) -- (-1.494,-1.933) -- (-1.321,-1.169) -- (-1.516,-0.902) -- (-1.746,-0.634) -- (-1.664,-0.269) -- (-1.706,0.047) -- (-1.814,0.397) -- (-2.089,0.892) -- (-1.429,0.960) -- (-0.990,0.979) -- (-0.770,1.118) -- (-0.524,1.190) -- (-0.225,0.974);
\fill[catS] (1.112,4.816) circle (1.15pt);
\fill[catS] (1.926,4.373) circle (1.15pt);
\fill[catS] (2.491,3.618) circle (1.15pt);
\fill[catS] (2.997,2.963) circle (1.15pt);
\fill[catS] (3.184,2.140) circle (1.15pt);
\fill[catS] (3.068,1.310) circle (1.15pt);
\fill[catS] (3.105,0.680) circle (1.15pt);
\fill[catS] (2.706,0.075) circle (1.15pt);
\fill[catS] (2.235,-0.362) circle (1.15pt);
\fill[catS] (2.692,-0.978) circle (1.15pt);
\fill[catS] (2.302,-1.370) circle (1.15pt);
\fill[catS] (1.818,-1.610) circle (1.15pt);
\fill[catS] (1.110,-1.435) circle (1.15pt);
\fill[catS] (1.155,-2.273) circle (1.15pt);
\fill[catS] (0.600,-2.072) circle (1.15pt);
\fill[catS] (0.148,-1.572) circle (1.15pt);
\fill[catS] (-0.197,-2.091) circle (1.15pt);
\fill[catS] (-0.527,-1.818) circle (1.15pt);
\fill[catS] (-0.951,-1.872) circle (1.15pt);
\fill[catS] (-1.494,-1.933) circle (1.15pt);
\fill[catS] (-1.321,-1.169) circle (1.15pt);
\fill[catS] (-1.516,-0.902) circle (1.15pt);
\fill[catS] (-1.746,-0.634) circle (1.15pt);
\fill[catS] (-1.664,-0.269) circle (1.15pt);
\fill[catS] (-1.706,0.047) circle (1.15pt);
\fill[catS] (-1.814,0.397) circle (1.15pt);
\fill[catS] (-2.089,0.892) circle (1.15pt);
\fill[catS] (-1.429,0.960) circle (1.15pt);
\fill[catS] (-0.990,0.979) circle (1.15pt);
\fill[catS] (-0.770,1.118) circle (1.15pt);
\fill[catS] (-0.524,1.190) circle (1.15pt);
\fill[catS] (-0.225,0.974) circle (1.15pt);
\draw[catP, line width=0.9pt] (-4.20,-8.35) -- (-3.50,-8.35);
\fill[catP] (-3.850,-8.350) circle (1.15pt);
\node[anchor=west, font=\sffamily\footnotesize] at (-3.40,-8.35) {Perception};
\draw[catR, line width=0.9pt] (-0.30,-8.35) -- (0.40,-8.35);
\fill[catR] (0.050,-8.350) circle (1.15pt);
\node[anchor=west, font=\sffamily\footnotesize] at (0.50,-8.35) {Reasoning};
\draw[catS, line width=0.9pt] (3.38,-8.35) -- (4.08,-8.35);
\fill[catS] (3.730,-8.350) circle (1.15pt);
\node[anchor=west, font=\sffamily\footnotesize] at (4.18,-8.35) {Insight};
\end{tikzpicture}}
\caption{\textbf{Capability-level scores of all 32 models on \bench{}} (accuracy, \%).
Each spoke represents a model, numbered and ordered clockwise from the top by overall accuracy.
The radial axis ranges from 20\% at the centre to 90\% at the rim. Exact scores are reported in \cref{tab:full}.}
\label{fig:polarall}
\end{figure}

\end{document}